\documentclass[11pt,x11names]{article}

\usepackage[T1]{fontenc}
\usepackage[utf8]{inputenc}
\usepackage[UKenglish]{isodate}

\usepackage{hyperref}

\usepackage{fancyhdr,cancel}

\usepackage{listings}
\usepackage[inline]{enumitem}
\usepackage{multicol,xcolor,booktabs}

\usepackage{times,soul}

\usepackage[hang,small,labelfont=bf,up,textfont=it,up]{caption}
\usepackage{subcaption}
\usepackage{amsmath,amsfonts,amsthm,amssymb,bm,dsfont}	
\setlist{nolistsep}

\usepackage{setspace}

\usepackage{graphicx}

\usepackage{adjustbox}

\usepackage[a4paper,top=3cm,bottom=3cm,left=2.25cm,right=2.25cm]{geometry}

\usepackage{centernot}

\usepackage{numprint}
\npthousandsep{,}\npthousandthpartsep{}\npdecimalsign{.}

\renewcommand{\vec}[1]{\boldsymbol{#1}}
\newcommand{\mvec}[1]{\mathbf{#1}}
\newcommand{\abs}[1]{\vert{#1}\vert}

\newcommand{\titledoc}{Latent structure blockmodels \\ for Bayesian spectral graph clustering}
\newcommand{\titleshort}{Latent structure blockmodels for Bayesian spectral graph clustering}

\providecommand{\keywords}[1]{{\small{\textbf{\textit{Keywords --- }} #1}}}

\usepackage{csquotes} 
\usepackage{natbib}

\usepackage{sectsty}
\partfont{\fontsize{17}{17}\selectfont}

\usepackage{thmtools}
\newtheorem{definition}{Definition}
\newtheorem{theorem}{Theorem}

\usepackage{authblk}

\usepackage[ruled,linesnumbered]{algorithm2e}
\SetKwInput{KwInput}{Input}
\hypersetup{colorlinks,linkcolor={blue},citecolor={blue},urlcolor={blue}}  
\numberwithin{equation}{section}

\usepackage{mathtools}
\mathtoolsset{showonlyrefs}

\author{Francesco Sanna Passino}
\author{Nicholas A. Heard}

\affil{Department of Mathematics, Imperial College London \\ 180 Queen's Gate, SW7 2AZ, London (United Kingdom)}

\date{}

\title{\huge\textbf{\titledoc}}

\counterwithout{equation}{section}

\theoremstyle{definition}
\newtheorem{example}{Example}

\begin{document}

\maketitle


\begin{abstract}
Spectral embedding of network adjacency matrices often produces node representations living approximately around low-dimensional submanifold structures. 
In particular, hidden substructure is expected to arise when the graph is generated from a latent position model.
Furthermore, the presence of communities within the network might generate community-specific submanifold structures in the embedding, but this is not explicitly accounted for in most statistical models for networks.
In this article, a class of models called latent structure block models (LSBM) is proposed to address such scenarios, allowing for graph clustering when community-specific one dimensional manifold structure is present. 
LSBMs focus on a specific class of latent space model, the random dot product graph (RDPG), and assign a latent submanifold to the latent positions of each community.
A Bayesian model for the embeddings arising from LSBMs is discussed, and shown to have a good performance on simulated and real world network data. 
The model is able to correctly recover the underlying communities living in a one-dimensional manifold, even when the parametric form of the underlying curves is unknown, achieving remarkable results on a variety of real data.
\end{abstract}


\section{Introduction} \label{intro}

Network-valued data are commonly observed in many real world applications. They are typically represented by graph adjacency matrices, consisting of binary indicators summarising which nodes are connected. 
Spectral embedding \citep{Luo03} is often the first preprocessing step in the analysis of graph adjacency matrices: the nodes are embedded onto a low-dimensional space via eigendecompositions or singular value decompositions. This work discusses a Bayesian network model for clustering the nodes of the graph in the embedding space, when community-specific substructure is present.
Therefore, the proposed methodology could be classified among spectral graph clustering methodologies, which have been extensively studied in the literature. Recent examples include \cite{Priebe19, Pensky19, Yang20, SannaPassino20}. More generally, spectral clustering is an active research area, extensively studied both from a theoretical and applied perspective \citep[for some examples, see][]{Couillet16, Hofmeyr19, Zhu19, Amini21, Han21}. In general, research efforts in spectral graph clustering \citep{Rohe11, Sussman12, Lyzinski14} have been primarily focused on studying its properties under simple community structures, for example the stochastic block model \citep[SBM,][]{Holland83}. On the other hand, in practical applications, it is common to observe more complex community-specific submanifold structure \citep{Priebe17,SannaPassino20}, which is not captured by spectral graph clustering models at present. In this work, a Bayesian model for spectral graph clustering under such a scenario is proposed, and applied on a variety of real-world networks. 

A popular statistical model for graph adjacency matrices is the latent position model \citep[LPM,][]{Hoff02}. Each node is assumed to have a low-dimensional vector representation $\vec x_i\in\mathbb R^d$, such that 
the probability of connection between two nodes is 
$\kappa(\vec x_i,\vec x_j)$, 
for a given \textit{kernel} function $\kappa:\mathbb R^d\times\mathbb R^d\to[0,1]$. Examples of commonly used kernel functions are the inner product $\kappa(\vec x_i,\vec x_j)=\vec x_i^\intercal\vec x_j$ \citep{Young07,Athreya18}, the logistic distance link $\kappa(\vec x_i,\vec x_j)=[1+\exp(-\|\vec x_i-\vec x_j\|)]^{-1}$ \citep{Hoff02,Salter17}, where $\|\cdot\|$ is a norm, or the Bernoulli-Poisson link $\kappa(\vec x_i,\vec x_j)=1-\exp(-\vec x_i^\intercal\vec x_j)$ \citep[for example,][]{Todeschini20}, where the latent positions 
lie in $\mathbb R_+^d$.
\citet{RubinDelanchy20} notes that spectral embeddings of adjacency matrices generated from LPMs produce node representations living near low-dimensional submanifold structures. 
 Therefore, for subsequent inferential tasks, for example clustering, it is necessary to employ methods which take into account this latent manifold structure.
 
\begin{figure}[!t]
\begin{minipage}{.43\textwidth}
\centering
\includegraphics[width=\textwidth]{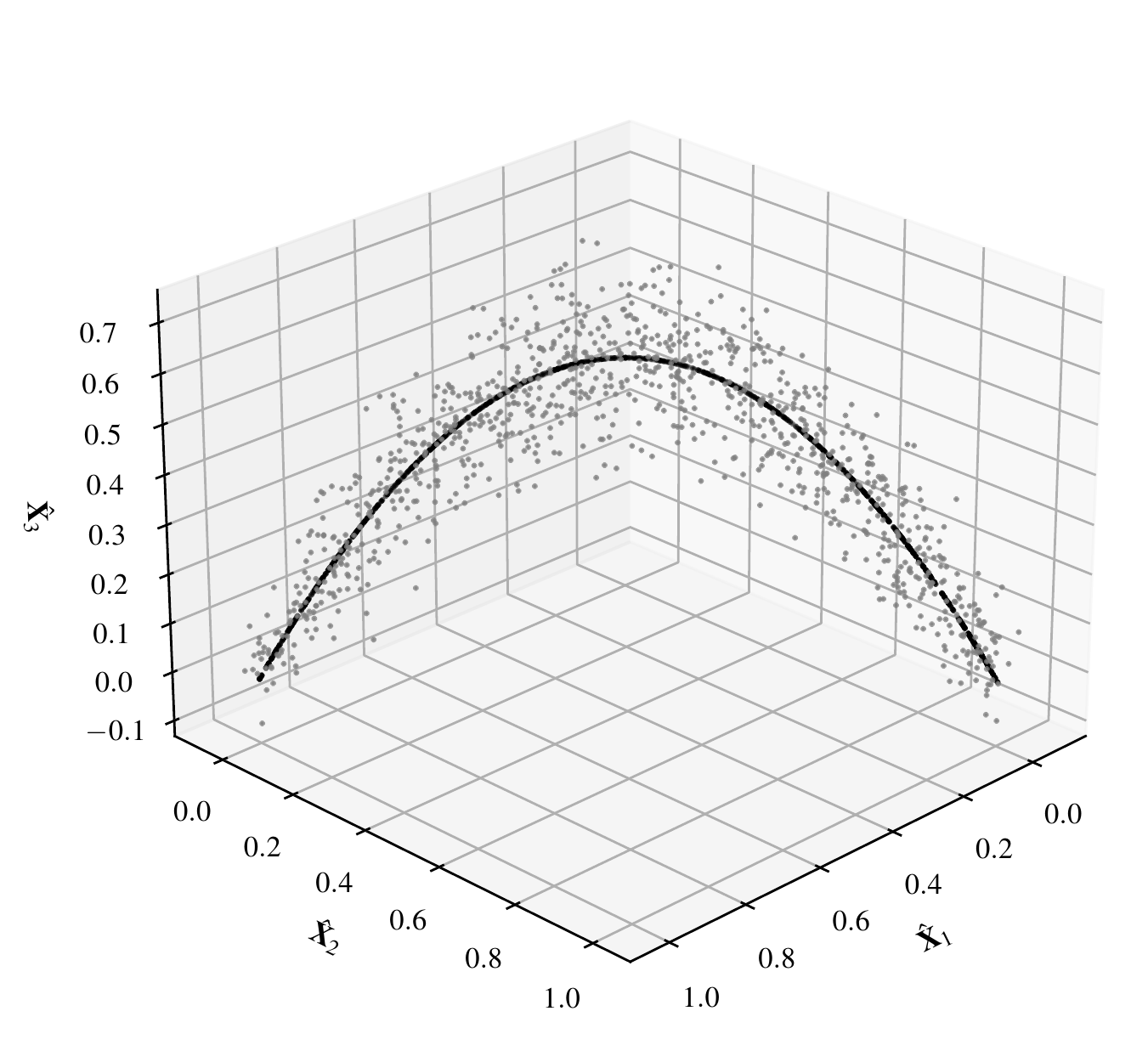}
\caption{Latent positions (\textbf{black}) and spectral estimates (\textbf{\color{gray} gray}) obtained from a simulated Hardy-Weinberg LSM.}
\label{hardy_ex}
\end{minipage}
\hspace{.02\textwidth}
\begin{minipage}{.53\textwidth}
\centering
\includegraphics[width=\textwidth]{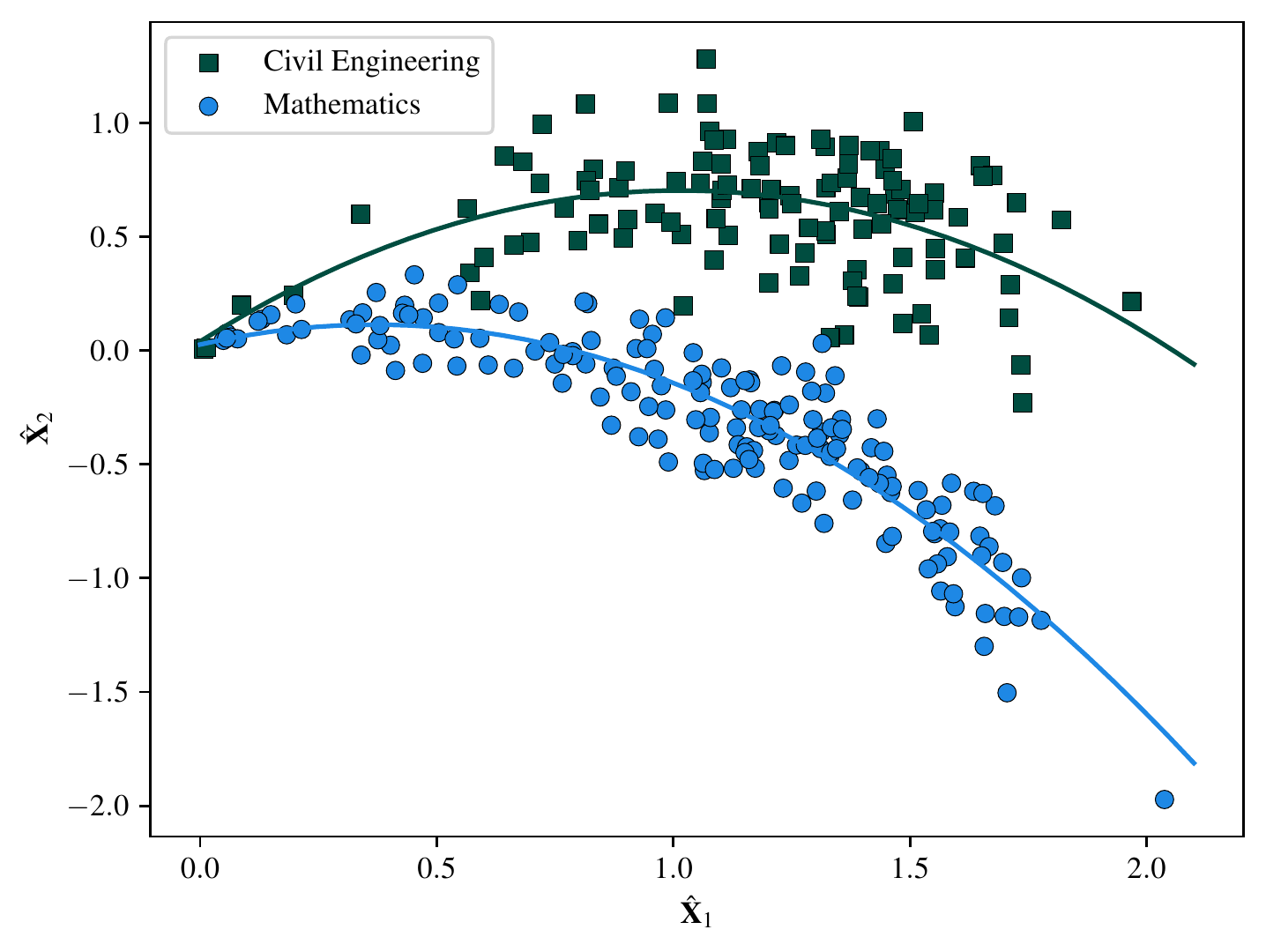}
\caption{Fitted quadratic curves for the two-dimensional (directed) spectral embedding of a computer network graph with two communities.}
\label{maths_cive}
\end{minipage}
\end{figure}

If the kernel function of the LPM is the inner product, spectral embedding provides consistent estimates of the latent positions \citep{Athreya16,RubinDelanchy17}, up to orthogonal rotations. This model is usually referred to as the random dot product graph \citep[RDPG,][]{Young07,Athreya18}.
In this work, RDPGs where the latent positions 
lie on a one-dimensional submanifold of the latent space are considered. Such a model is known in the literature as a latent structure model \citep[LSM,][]{Athreya21}. %
An example is displayed in Figure~\ref{hardy_ex}, which shows the latent positions, and corresponding estimates obtained using adjacency spectral embedding (ASE; see Section~\ref{sec:rdpg} for details), of a simulated graph with $1000$ nodes, where the latent positions are assumed to be drawn from the Hardy-Weinberg curve $\vec f(\theta)=\{\theta^2,(1-\theta)^2,2\theta(1-\theta)\}$, with $\theta\sim\text{Uniform}[0,1]$ \citep{Athreya21,Trosset20}.

In addition, nodes are commonly divided into inferred groups 
such that it can be plausible to assume that the latent positions live on group-specific submanifolds. The objective of community detection on graphs is to recover such latent groups. 
For example, in computer networks, it is common to observe community-specific curves in the latent space. Figure~\ref{maths_cive} displays an example, representing the two-dimensional latent positions, estimated using directed adjacency spectral embedding (DASE; see Section~\ref{sec:rdpg}), of $278$ computers in a network of HTTP/HTTPS connections from machines hosted in two computer laboratories at Imperial College London (further details are given in Section~\ref{sec:icl}). The latent positions corresponding to the two communities appear to be distributed around two quadratic curves with different parameters. The best fitting quadratic curves in $\mathbb R^2$ passing through the origin are also displayed, representing an estimate of the underlying community-specific submanifolds.

Motivated by the practical application to computer networks, this article develops inferential methods for LSMs allowing for latent community structure. Nodes belonging to different communities are assumed to correspond to community-specific submanifolds. This leads to the definition of \textit{latent structure blockmodels (LSBM)}, which admit communities with manifold structure in the latent space. The main contribution is a Bayesian probability model for LSBM embeddings, which enables community detection in RDPG models with underlying community-specific manifold structure.

Naturally, for clustering purposes, procedures that exploit the structure of the submanifold are expected to perform more effectively than standard alternatives such as spectral clustering with $k$-means \citep{vonLuxburg,Rohe11} or Gaussian mixture models \citep{RubinDelanchy17}. 
The problem of clustering in the presence of group-specific curves has been addressed for distance-based clustering methods \citep{Diday71,Diday76}, often using kernel-based methodologies \citep[see \cite{Bouveyron15}, and the review of][]{Smola18} or, more recently, deep learning techniques \citep{Ye20}, but it is understudied for the purposes of graph clustering.
Exploiting the underlying structure is often impractical, since the submanifold is usually unknown. Within the context of LSMs, \citet{Trosset20} propose using an extension of Isomap \citep{Tenenbaum00} in the presence of noise to learn a one-dimensional unknown manifold from the estimated latent positions. In this work, the problem of submanifold estimation is addressed using flexible functional models, namely Gaussian processes, in the embedding space. Applications on simulated and real world computer network data show that the proposed methodology is able to successfully recover the underlying communities even in complex cases with substantial overlap between the groups.
Recent work from \cite{Dunson21} 
also demonstrate the potential of Gaussian processes for manifold inference from noisy observations. 

Geometries arising from network models have been extensively studied in the literature \citep[see, for example,][]{Asta15,McCormick15,Smith19}.
The approach in this paper is also related to the findings of \cite{Priebe17}, where a semiparametric Gaussian mixture model with quadratic structure is used to model the embeddings arising from a subset of the neurons in the \textit{Drosophila} connectome. In particular, the \textit{Kenyon Cell} neurons are not well represented by a stochastic blockmodel, which is otherwise appropriate for other neuron types. Consequently, those neurons are modelled via a continuous curve in the latent space, estimated via semiparametric maximum likelihood \citep{Kiefer56,Lindsay83}. 
However, the real world examples presented in this article in Section~\ref{sec:results} require community-specific curves in the latent space for all the communities, not only for a subset of the nodes. Furthermore, estimation can be conducted within the Bayesian paradigm, which conveniently allows marginalisation of the model parameters.

This article is organised as follows: Section~\ref{sec:rdpg} formally introduces RDPGs and spectral embedding methods. Sections~\ref{sec:lsbm} and~\ref{sec:bayes_lsbm} present the main contributions of this work: the latent structure blockmodel (LSBM), and Bayesian inference for the communities of LSBM embeddings. The method is applied to simulated and real world networks in Section~\ref{sec:results}, and the article is then concluded with a discussion in Section~\ref{sec:conclusion}. 

\section{Random dot product graphs and latent structure models} \label{sec:rdpg}

Mathematically, a network $\mathbb G=(V,E)$ is represented by a set $V=\{1,2,\dots,n\}$ of nodes, and an edge set $E$, such that $(i,j)\in E$ if $i$ forms a link with $j$. If $(i,j)\in E \Rightarrow (j,i)\in E$, the network is \textit{undirected}, otherwise it is \textit{directed}. 
If the node set is partitioned into two sets $V_1$ and $V_2$, $V_1\cap V_2=\varnothing$, such that $E\subset V_1\times V_2$, the graph is \textit{bipartite}.
A network can be summarised by the adjacency matrix $\mvec A\in\{0,1\}^{n\times n}$, such that $A_{ij}=\mathds 1_E\{(i,j)\}$, where $\mathds 1_\cdot\{\cdot\}$ is an indicator function. Latent position models \citep[LPMs,][]{Hoff02} for undirected graphs postulate that the edges are sampled independently, with 
\begin{equation}
\mathbb P(A_{ij}=1\mid\vec x_i,\vec x_j)=\kappa(\vec x_i,\vec x_j). 
\end{equation}
A special case of LPMs is the random dot product graph \citep[RDPG,][]{Young07}, defined below.

\begin{definition}[Random dot product graph] 
For an integer $d>0$, let $F$ be an inner product distribution on $\mathcal X\subset \mathbb R^d$, such that for all $\vec x,\vec x^\prime\in\mathcal X$, $0\leq \vec x^\intercal\vec x^\prime\leq 1$. Also, let $\mvec{X}=(\vec x_1,\dots,\vec x_n)^\intercal\in\mathcal{X}^n$ and $\mvec A\in\{0,1\}^{n\times n}$ be symmetric. Then $(\mvec A,\mvec X)\sim\text{RDPG}_d(F^n)$ if
\begin{align}
  \vec x_1,\dots,\vec x_n &\overset{iid}\sim F\\
  A_{ij}\vert \mvec X &\sim \textnormal{Bernoulli}(\vec x_i^\intercal\vec x_j),\quad 1\leq i<j\leq n. \label{rdpg}
\end{align}
\end{definition}

Given a realisation of the adjacency matrix $\mvec A$, the first inferential objective is to estimate the latent positions $\vec x_1,\dots, \vec x_n$. In RDPGs, the latent positions are inherently unidentifiable; in particular, multiplying the latent positions by any orthogonal matrix $\mvec Q\in\mathbb O(d)$, the orthogonal group with signature $d$, leaves the inner product \eqref{rdpg} unchanged: $\vec x^\intercal\vec x^\prime=(\mvec Q\vec x)^\intercal\mvec Q\vec x^\prime$. Therefore, the latent positions can only be estimated \textit{up to orthogonal rotations}. Under such a restriction, the latent positions are consistently estimated via spectral embedding methods. In particular, the adjacency spectral embedding (ASE), defined below, has convenient asymptotic properties.

\begin{definition}[ASE -- Adjacency spectral embedding] \label{adj_emb}
For a given integer $d\in\{1,\ldots,n\}$ and a binary symmetric adjacency matrix $\mvec A\in\{0,1\}^{n\times n}$, the $d$-dimensional adjacency spectral embedding (ASE) $\hat{\mvec X}=[\hat{\vec x}_{1},\dots,\hat{\vec x}_{n}]^\intercal$ of $\mvec A$ is 
\begin{equation}
\hat{\mvec X} = \mvec\Gamma\mvec\Lambda^{1/2}\in\mathbb R^{n\times d},
\end{equation}
where $\mvec\Lambda$ is a $d\times d$ diagonal matrix containing the absolute values of the $d$ largest eigenvalues in magnitude, in decreasing order, and $\mvec\Gamma$ is a $n\times d$ matrix containing corresponding orthonormal eigenvectors.
\end{definition}

Alternatively, the Laplacian spectral embedding is also frequently used, and considers a spectral decomposition of the modified Laplacian $\mvec D^{-1/2}\mvec A\mvec D^{-1/2}$ instead, where $\mvec D$ is the degree matrix. In this work, the focus will be mainly on ASE. 
Using ASE, the latent positions are consistently estimated up to orthogonal rotations, and a central limit theorem is available \citep{Athreya16,RubinDelanchy17,Athreya18}. 

\begin{theorem}[ASE-CLT -- ASE central limit theorem]
\label{clt}
Let $(\mvec A^{(n)}, \mvec X^{(n)}) \sim\text{RDPG}_d(F^n), n=1,2,\dots$, be a sequence of adjacency matrices and corresponding latent positions, and let $\hat{\mvec X}^{(n)}$ be the $d$-dimensional ASE of $\mvec A^{(n)}$. For an integer $m>0$, and for the sequences of points $\vec x_1,\dots,\vec x_m\in\mathcal X$ and $\vec u_1,\dots,\vec u_m\in\mathbb R^d$, there exists a sequence of orthogonal matrices $\mvec Q_1,\mvec Q_2,\ldots\in\mathbb O(d)$ such that:
\begin{equation}
\lim_{n\to\infty}\mathbb P\left\{\bigcap_{i=1}^m \sqrt n\left(\mvec Q_n\hat{\vec x}_i^{(n)} - \vec x_i^{(n)}\right) \leq \vec u_i\ \Bigg\vert\ 
\vec x_i^{(n)}=\vec x_i
\right\} = \prod_{i=1}^m \Phi\{\vec u_i,\bm\Sigma(\vec x_i)\},
\end{equation}
where $\Phi\{\cdot\}$ is the CDF of a $d$-dimensional normal distribution, and $\bm\Sigma(\cdot)$ is a covariance matrix which depends on the true value of the latent position. 
\end{theorem}

Therefore, informally, it could be assumed that, for $n$ large,
$\mvec Q\hat{\vec x}_i\sim\mathbb N_d\left\{\vec x_i,\bm\Sigma(\vec x_i)\right\}$ 
for some $\mvec Q\in\mathbb O(d)$, where $\mathbb N_d\{\cdot\}$ is the $d$-dimensional multivariate normal distribution.
Theorem~\ref{clt} provides strong theoretical justification for the choice of normal likelihood assumed in the Bayesian model presented in the next section. 

This article is mainly concerned with community detection in RDPGs when the latent positions 
lie on community-specific one-dimensional submanifolds $\mathcal S_k\subset\mathbb R^d$, $k=1,\dots,K$.
The proposed methodology builds upon latent structure models \citep[LSMs,][]{Athreya21}, a 
subset of RDPGs.
In LSMs, it is assumed that the latent positions of the nodes are determined by draws from a univariate underlying distribution $G$ on $[0,1]$, inducing a distribution $F$ on a structural support univariate submanifold $\mathcal S\subset\mathbb R^d$, such that:
\begin{equation}
\vec x_i \overset{iid}{\sim} F, 
\ i=1,\dots,n.
\label{lsm_generic}
\end{equation}
In particular, the distribution $F$ on $\mathcal S$ is the distribution of the vector-valued transformation $\vec f(\theta)$ of a univariate random variable $\theta\sim G$, where $\vec f:[0,1]\to\mathcal S$ is a function mapping draws from $\theta$ to $\mathcal S$. The inverse function $\vec f^{-1}:\mathcal S\to[0,1]$ could be interpreted as the arc-length parametrisation of $\mathcal S$. In simple terms, each node is assigned a draw $\theta_i$ from the underlying distribution $G$, representing \textit{how far} along the submanifold $\mathcal S$ the corresponding latent position lies, such that:
\begin{equation}
\vec x_i=\vec f(\theta_i).
\end{equation} 
Therefore, conditional on $\vec f$, the latent positions are uniquely determined by $\theta_i$.

If the graph is directed or bipartite, each node is assigned two latent positions $\vec x_i$ and $\vec x_i^\prime$, and the random dot product graph model takes the form $\mathbb P(A_{ij}=1\mid\vec x_i,\vec x_j^\prime)=\vec x_i^\intercal\vec x_j^\prime$. In this case, the singular value decomposition of $\mvec A$ can be used over the eigendecomposition. 

\begin{definition}[DASE -- Directed adjacency spectral embedding] \label{dase}
For an 
integer $d\in\{1,\ldots,n\}$ and an 
adjacency matrix $\mvec A\in\{0,1\}^{n\times n}$, the $d$-dimensional directed adjacency spectral embeddings (DASE) $\hat{\mvec X}=[\hat{\vec x}_{1},\dots,\hat{\vec x}_{n}]^\intercal$ and $\hat{\mvec X}^\prime=[\hat{\vec x}_{1}^\prime,\dots,\hat{\vec x}_{n}^\prime]^\intercal$ of $\mvec A$ are
\begin{equation}
\hat{\mvec X} =  \mvec U\mvec S^{1/2} \in\mathbb R^{n\times d},\ \hat{\mvec X}^\prime = \mvec V\mvec S^{1/2}\in\mathbb R^{n\times d},
\end{equation}
where $\mvec S$ is a $d\times d$ diagonal matrix containing the $d$ largest singular values, in decreasing order, and $\mvec U$ and $\mvec V$ are a $n\times d$ matrix containing corresponding left- and right-singular vectors.
\end{definition}

Note that ASE and DASE could also be interpreted as instances of multidimensional scaling \citep[MDS,][]{Torgerson52, Shepard62}.
Clearly, DASE could be also used on rectangular adjacency matrices arising from bipartite graphs. Furthermore, \citet{Jones20} prove the equivalent of Theorem~\ref{clt} for DASE, demonstrating that DASE also provides a consistent procedure for estimation of the latent positions in directed or bipartite graphs. Analogous constructions to undirected LSMs could be posited for directed or bipartite models, where the $\vec x_i$'s and $\vec x_i^\prime$'s lie on univariate structural support submanifolds.

\section{Latent structure blockmodels} \label{sec:lsbm}

This section extends LSM, explicitly allowing for community structure. In particular, it is assumed that each node is assigned a latent community membership $z_i\in\{1,\dots,K\},\ i=1,\dots,n$, and each community is associated with a different one-dimensional structural support submanifold $\mathcal S_k\subset \mathbb R^d,\ k=1,\dots,K$. Therefore, it is assumed that $F=\sum_{k=1}^K \eta_kF_k$ is a mixture distribution with components $F_1,\dots,F_K$ supported on the submanifolds $\mathcal S_1,\dots,\mathcal S_K$, with weights $\eta_1,\dots,\eta_K$ such that for each $k$, $\eta_k\geq0$ and $\sum_{k=1}^K\eta_k=1$. Assuming community allocations $\vec z=(z_1,\dots,z_n)$, the latent positions are obtained as 
\begin{equation}
\vec x_i\mid z_i \sim F_{z_i},\ i=1,\dots,n,
\end{equation}
where, similarly to \eqref{lsm_generic}, $F_{z_i}$ is the distribution of the community-specific vector-valued transformation $\vec f_{z_i}(\theta)$, of a univariate random variable $\theta\sim G$, which is instead shared across communities. The vector-valued functions $\vec f_k=(f_{k,1},\dots,f_{k,d}):\mathcal G\to\mathcal S_k,\ k=1,\dots,K,$ map the latent draw from the distribution $G$ to $\mathcal S_k$. 
The resulting model will be referred to as the \textit{latent structure blockmodel} (LSBM). Note that, for generality, the support of the underlying distribution $G$ is assumed here to be $\mathcal G\subset\mathbb R$. Furthermore, $G$ is common to \textit{all} the nodes, and the pair $(\theta_i,z_i)$, where $\theta_i\sim G$, uniquely determines the latent position $\vec x_i$ through $\vec f_{z_i}$, such that
\begin{equation}
\vec x_i=\vec f_{z_i}(\theta_i).
\end{equation}
Note that, 
in the framework described above, the 
submanifolds $\mathcal S_1,\dots,\mathcal S_K$ are one-dimensional, corresponding to curves, but the LSBM could be extended to higher-dimensional settings of the underlying subspaces, postulating 
draws from a multivariate distribution $G$ supported on $\mathcal G\subseteq \mathbb R^p$ with $1\leq p<d$. 


Since the latent structure blockmodel is a special case of the random dot product graph, the LSBM latent positions are also estimated consistently via ASE, up to orthogonal rotations, conditional on knowing the functions $\vec f_k(\cdot)$. Therefore, by Theorem~\ref{clt}, approximately:
\begin{equation}
\mvec Q\hat{\vec x}_i\sim\mathbb N\{\vec f_{z_i}(\theta_i),\bm\Sigma(\vec x_i)\}. 
\label{normality}
\end{equation}

Special cases of the LSBM include 
the stochastic blockmodel \citep[SBM,][]{Holland83}, 
the degree-corrected 
SBM \citep[DCSBM,][]{Karrer11}, and other more complex latent structure models with clustering structure, as demonstrated in the following examples.
Note that, despite the similarity in name, LSBMs
are \emph{different} from latent block models \citep[LBMs; see, for example,][]{Keribin15,Wyse17}, which instead generalise the SBM to bipartite graphs. 

\begin{example}[Stochastic blockmodel, SBM] \label{sbm_sim}
In an SBM \citep{Holland83}, the edge probability is determined by the community allocation of the nodes: $A_{ij}\sim\textnormal{Bernoulli}(B_{z_i z_j})$, where $\mvec B\in[0,1]^{K\times K}$ is a matrix of probabilities for connections between communities. An SBM characterised by a non-negative definite matrix $\mvec B$ of rank $d$ can be expressed as an LSBM, assigning community-specific latent positions $\vec x_i=\vec\nu_{z_i}\in\mathbb R^d$, such that for each $(k,\ell)$, $B_{k\ell}=\vec\nu_k^\intercal\vec\nu_\ell$. Therefore, $\vec f_k(\theta_i)=\vec\nu_k$, with each $\theta_i=1$ for identifiability. It follows that $f_{k,j}(\theta_i)=\nu_{k,j}$. 
\end{example}

\begin{example}[Degree-corrected SBM, DCSBM] \label{dcsbm_sim}
DCSBMs \citep{Karrer11} extend SBMs, allowing for heterogeneous degree distributions within communities. The edge probability depends on the community allocation of the nodes, and degree-correction parameters $\theta\in\mathbb R^n$ for each node, such that $A_{ij}\sim\textnormal{Bernoulli}(\theta_i\theta_jB_{z_i z_j})$. In a latent structure blockmodel interpretation, the latent positions are $\vec x_i=\theta_i\vec\nu_{z_i}\in\mathbb R^d$. Therefore, $\vec f_k(\theta_i)=\theta_i\vec\nu_k$, with 
$f_{k,j}(\theta_i)=\theta_i\nu_{k,j}$. 
For identifiability, one could set each $\nu_{k,1}=1$.
\end{example}

\begin{example}[Quadratic latent structure blockmodel] \label{quad_sim}
For an LSBM with quadratic $\vec f_k(\cdot)$, it can be postulated that, conditional on a community allocation $z_i$, $\vec x_i=\vec f_{z_i}(\theta_i)=\bm\alpha_{z_i}\theta_i^2+\bm\beta_{z_i}\theta_i+\bm\gamma_{z_i}$, with $\bm\alpha_k,\bm\beta_k,\bm\gamma_k\in\mathbb R^d$. 
Under this model: $f_{k,j}(\theta_i)=\alpha_{k,j}\theta_i^2+\beta_{k,j}\theta+\gamma_{k,j}$. 
Note that the model is not identifiable: for $v\in\mathbb R$, then $ (\bm\alpha_{z_i}/v^2)(v\theta_i)^2+(\bm\beta_{z_i}/v)(v\theta_i)+\bm\gamma_{z_i}$ is equivalent to $\vec f_{z_i}(\theta_i)$. A possible solution is to fix each $\beta_{k,1}=1$.
\end{example}

\begin{figure*}[t]
\centering
\begin{subfigure}[t]{0.325\textwidth}
\centering
\caption{SBM} 
\includegraphics[width=\textwidth]{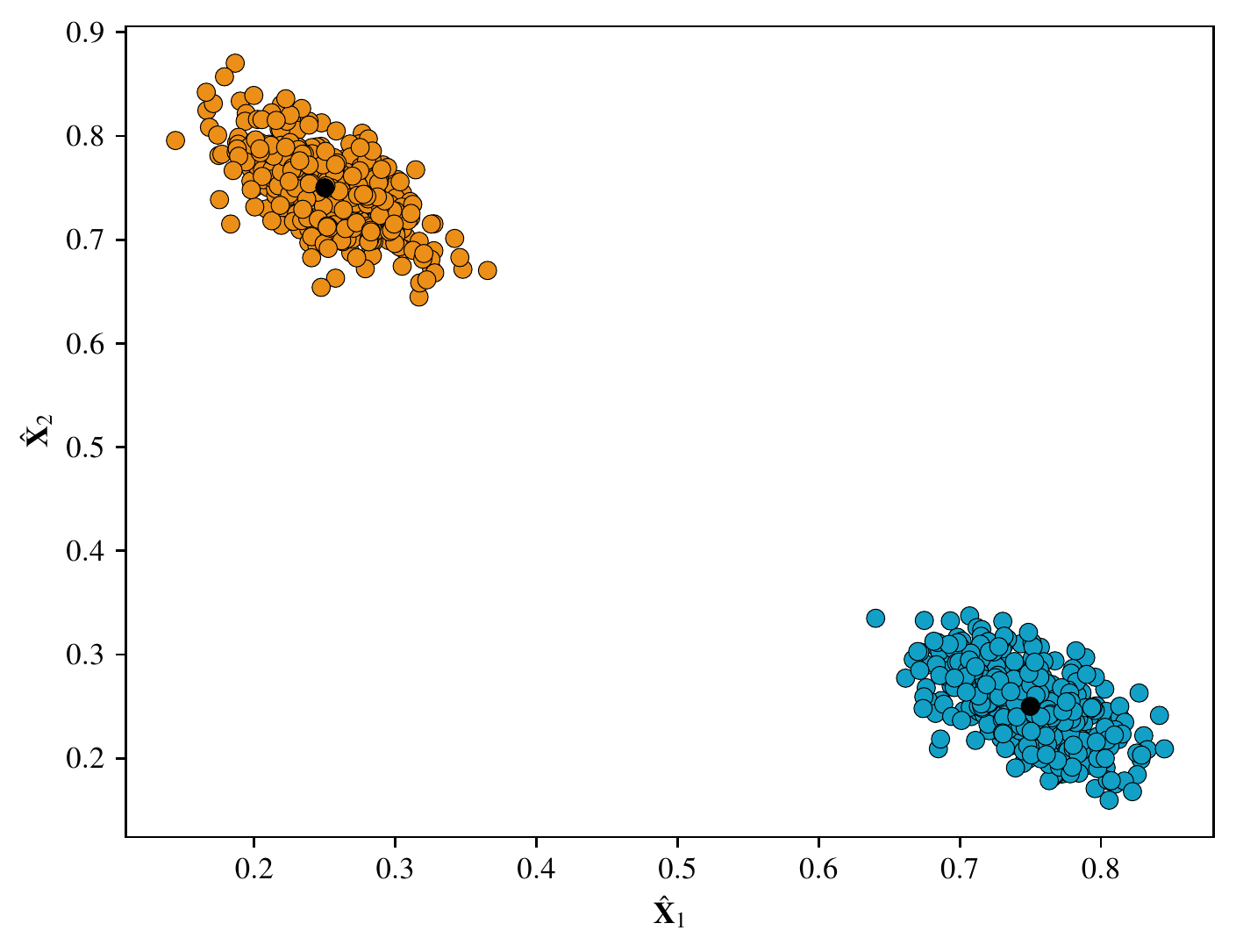}
\end{subfigure}
\begin{subfigure}[t]{0.325\textwidth}
\centering
\caption{DCSBM}
\includegraphics[width=\textwidth]{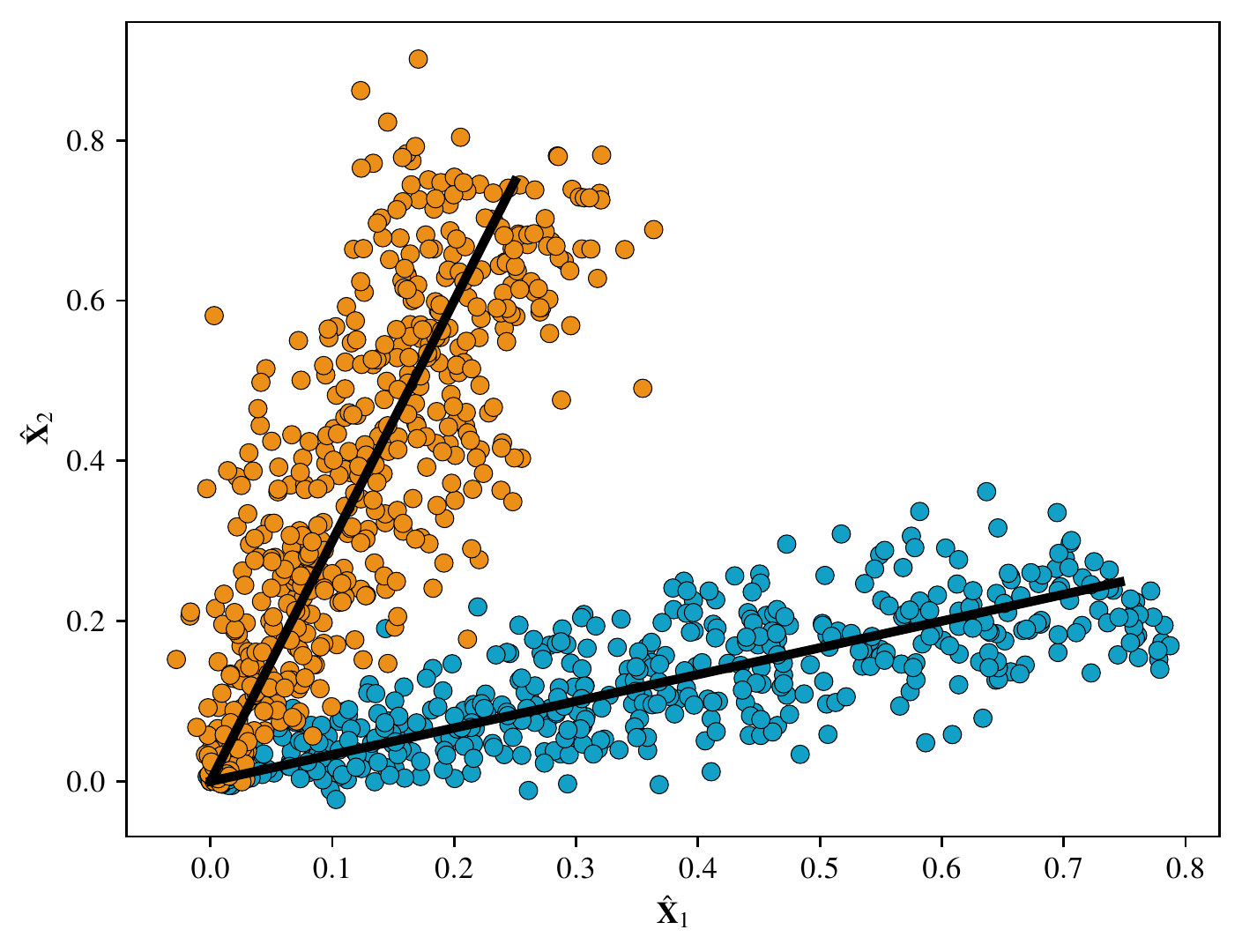}
\end{subfigure}
\begin{subfigure}[t]{0.325\textwidth}
\centering
\caption{Quadratic model}
\includegraphics[width=\textwidth]{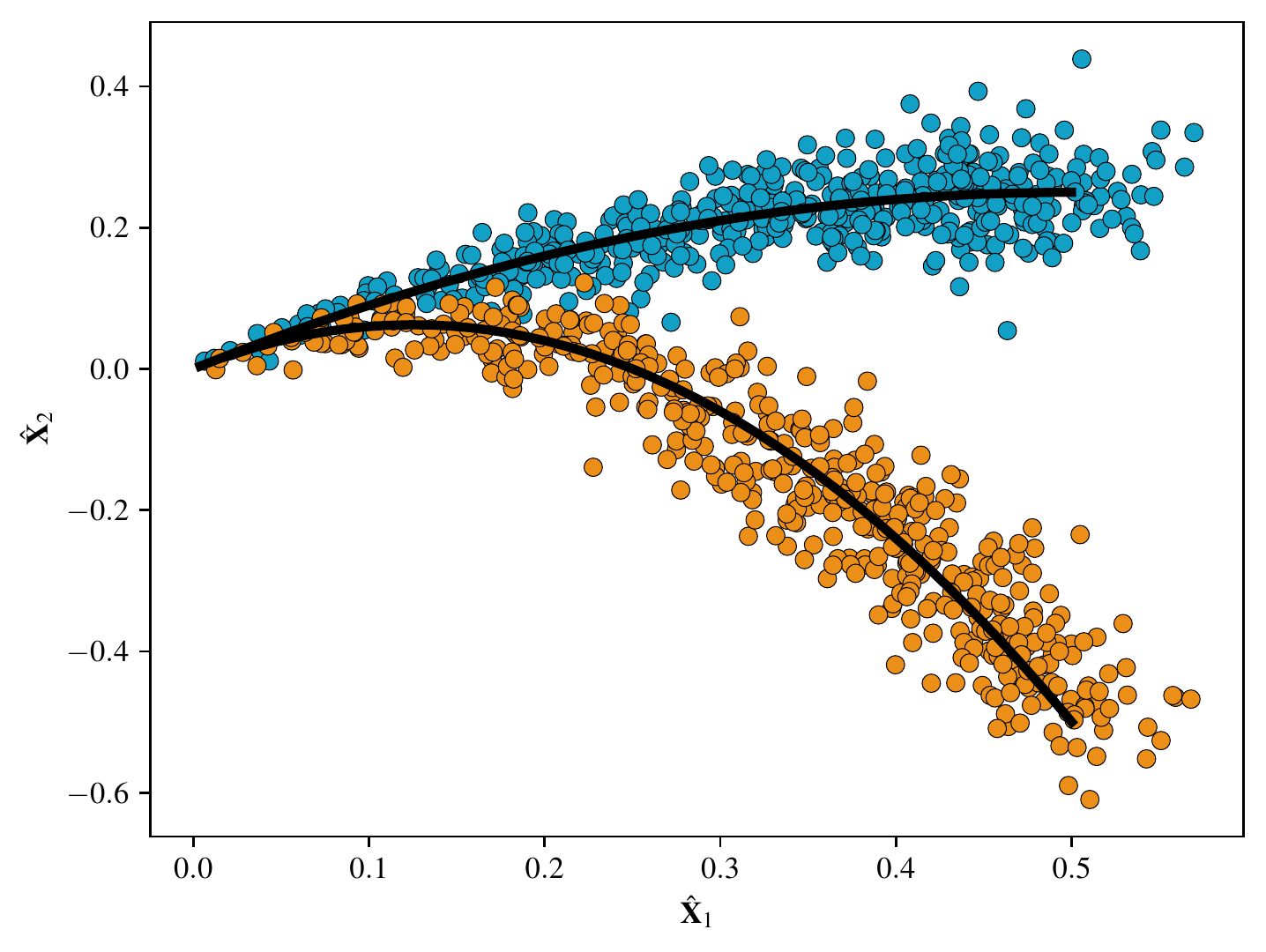}
\label{quad_plot}
\end{subfigure}
\caption{Scatterplots of the two-dimensional ASE of simulated graphs arising from the models in Examples~\ref{sbm_sim}, \ref{dcsbm_sim}, \ref{quad_sim}, and true underlying latent curves (in black). For each graph, $n=1000$ with $K=2$ communities of equal size. For {(a)} and {(b)}, $\bm\nu_1=[3/4,1/4]$, $\bm\nu_2=[1/4,3/4]$ and $\theta_i\sim\text{Beta}(1,1)$, \textit{cf.} Examples~\ref{sbm_sim} and \ref{dcsbm_sim}. For {(c)}, $\bm\alpha_k=[-1, -4]$, $\bm\beta_k=[1,1]$, $\bm\gamma_k=[0,0]$ and $\theta_i\sim\text{Beta}(2,1)$, \textit{cf.} Example~\ref{quad_sim}.}
\label{sim_graphs}
\end{figure*}

Figure~\ref{sim_graphs} shows example embeddings arising from the three models described above. From these plots, it is evident that taking into account the underlying structure is essential for successful community detection.

\section{Bayesian modelling of LSBM embeddings} \label{sec:bayes_lsbm}

Under the LSBM, the inferential objective is to recover the community allocations $\vec z=(z_1,\dots,z_n)$ given a realisation of the adjacency matrix $\mvec A$. Assuming normality of the rows of ASE for LSBMs \eqref{normality}, the inferential problem consists of making joint inference about $\vec z$ and the 
latent functions $\vec f_k=(f_{k,1},\dots,f_{k,d}):\mathcal G\to\mathbb R^d$. The prior 
for $\vec z$ follows a Categorical-Dirichlet structure: 
\begin{align}
z_i &\sim \text{Categorical}(\bm\eta),\ \bm\eta=(\eta_1,\dots,\eta_K),\ i=1,\dots,n, \\ 
\bm\eta&\sim \text{Dirichlet}(\nu/K,\dots,\nu/K), \label{prior_structure}
\end{align}
where $\nu,\eta_k\in\mathbb R_+,\ k\in\{1,\dots,K\}$, and $\sum_{k=1}^K\eta_k=1$.

Following the ASE-CLT in Theorem~\ref{clt}, the estimated latent positions are assumed to be drawn from Gaussian distributions centred at the underlying function value. 
Conditional on the pair $(\theta_i,z_i)$, the following distribution is postulated for $\hat{\vec x}_i$: 
\begin{equation}
\hat{\vec x}_i\ \vert\ \theta_i,\vec f_{z_i},\bm\sigma^2_{z_i} \sim \mathbb N_d\left\{\vec f_{z_i}(\theta_i),\bm\sigma^2_{z_i}\mvec I_d\right\},\ i=1,\dots,n,\label{eq:xhat}
\end{equation}
where $\bm\sigma^2_k=(\sigma^2_{k,1},\dots,\sigma^2_{k,d})\in\mathbb R_+^d$ is a community-specific vector of variances and $\mvec I_d$ is the $d\times d$ identity matrix. 
Note that, for simplicity, the components of the estimated latent positions are assumed to be independent. 
This assumption loosely corresponds to the $k$-means clustering approach, which has been successfully deployed in spectral graph clustering under the SBM \citep{Rohe11}. Here, the same idea is extended to a functional setting. Furthermore, for tractability \eqref{eq:xhat} assumes the variance of $\hat{\vec x}_i$ does not depend on $\vec x_i$, but only on the community allocation $z_i$. 

For a full Bayesian model specification, prior distributions are required for the latent functions and the 
variances. The most popular prior for 
functions is the Gaussian process \citep[GP; see, for example,][]{Rasmussen06}. 
Here, for each community $k$, the $j$-th dimension of the true latent positions are assumed to lie on a one-dimensional manifold described by a function $f_{k,j}$ with a hierarchical GP-IG prior, with an inverse gamma (IG) prior on the variance:
\begin{align}
f_{k,j} \vert \sigma^2_{k,j} &\sim \text{GP}(0,\sigma^2_{k,j}\xi_{k,j}),\ k=1,\dots,K,\ j=1,\dots,d, \notag \\
\sigma^2_{k,j} &\sim \text{IG}(a_0,b_0),\ k=1,\dots,K,\ j=1,\dots,d, 
\label{model_gp}
\end{align}
where $\xi_{k,j}(\cdot,\cdot)$ is a positive semi-definite kernel function and $a_0,b_0\in\mathbb R_+$. 
Note that the terminology \textquote{kernel} is used in the literature for both the GP covariance function $\xi_{k,j}(\cdot,\cdot)$ and the function $\kappa(\cdot,\cdot)$ used in LPMs (\textit{cf.} Section~\ref{intro}), but their meaning is \textit{fundamentally different}. In particular, $\kappa:\mathbb R^d\times\mathbb R^d\to[0,1]$ is a component of the \textit{graph generating process}, and assumed here to be the inner product, corresponding to RDPGs. On the other hand, $\xi_{k,j}:\mathbb R\times\mathbb R\to\mathbb R$ is the scaled covariance function of the GP prior on the unknown function $f_{k,j}$, which is used for modelling the observed graph embeddings, or equivalently the \textit{embedding generating process}. There are no restrictions on the possible forms of $\xi_{k,j}$, except positive semi-definiteness. Overall, the approach is similar to the overlapping mixture of Gaussian processes method \citep{Lazaro12}.

The class of models that can be expressed in the form \eqref{model_gp} is vast, and includes, for example, polynomial regression and splines, under a conjugate normal-inverse-gamma prior for the regression coefficients. For example, consider any function that can be expressed in the form $f_{z_i,j}(\theta_i)= \bm\phi_{z_i,j}(\theta_i)^\intercal\vec w_{z_i,j}$ for some community-specific basis functions $\bm\phi_{k,j}:\mathbb R\to\mathbb R^{q_{k,j}}, q_{k,j}\in\mathbb Z_+,$ and corresponding coefficients $\vec w_{k,j}\in\mathbb R^{q_{k,j}}$. If the coefficients are given a normal-inverse-gamma prior
\begin{align}
  (\vec w_{k,j},\sigma^2_{k,j})&\sim\text{NIG}(\vec 0,\bm\Delta_{k,j},a_0,b_0) =\mathbb N_{q_{k,j}}(\vec 0,\sigma^2_{k,j}\bm\Delta_{k,j})\text{IG}(a_0,b_0),
\end{align}
where $\bm\Delta_{k,j}\in\mathbb R^{q_{k,j}\times q_{k,j}}$ is a positive definite matrix, then $f_{k,j}$ takes the form \eqref{model_gp}, with the kernel function
\begin{equation}
\xi_{k,j}(\theta,\theta^\prime) = \bm\phi_{k,j}(\theta)^\intercal\bm\Delta_{k,j}\bm\phi_{k,j}(\theta^\prime). \label{dot_kernel}
\end{equation} 
Considering the examples in Section~\ref{sec:lsbm}, the SBM (\textit{cf.} Example~\ref{sbm_sim}) corresponds to $\xi_{k,j}(\theta,\theta^\prime)=\Delta_{k,j},\ \Delta_{k,j}\in\mathbb R_+$, whereas the DCSBM (\textit{cf.} Example~\ref{dcsbm_sim}) corresponds to $\xi_{k,j}(\theta,\theta^\prime)=\theta\theta^\prime\Delta_{k,j},\ \Delta_{k,j}\in\mathbb R_+$. For the quadratic LSBM (\textit{cf.} Example~\ref{quad_sim}), the GP kernel takes the form $\xi_{k,j}(\theta,\theta^\prime)=(1,\theta,\theta^2)\bm\Delta_{k,j}(1,\theta^\prime,\theta^{\prime2})^\intercal$ for a positive definite scaling matrix $\bm\Delta_{k,j}\in\mathbb R^{3\times 3}$.

The LSBM specification is completed with a prior for each $\theta_i$ value, which specifies the unobserved location of the latent position $\vec x_i$ along each submanifold curve; for $\mu_\theta\in\mathbb R$, $\sigma^2_\theta\in\mathbb R_+$,
\begin{equation}
  \theta_i \sim \mathbb N(\mu_\theta,\sigma^2_\theta),\ i=1,\dots,n. 
\label{model}
\end{equation} 

\subsection{Posterior and marginal distributions}

The posterior distribution for $(f_{k,j},\sigma^2_{k,j})$ has the same GP-IG structure as \eqref{model_gp}, with updated parameters:
\begin{align}
f_{k,j} \vert \sigma^2_{k,j}, \vec z, \bm\theta, \hat{\mvec X} &\sim \text{GP}(\mu_{k,j}^\star,\sigma^2_{k,j}\xi_{k,j}^\star), \\ 
\sigma^2_{k,j} \vert \vec z, \bm\theta, \hat{\mvec X} &\sim \text{Inv-Gamma}(a_k,b_{k,j}), 
\end{align}
with $k=1,\dots,K,\ j=1,\dots,d$.
The parameters are updated as follows:
\begin{align}
&\mu_{k,j}^\star(\theta) = \bm\Xi_{k,j}(\theta, \bm\theta_k^\star)\{\bm\Xi_{k,j}(\bm\theta_k^\star, \bm\theta_k^\star) + \mvec I_{n_k}\}^{-1}\hat{\mvec X}_{k,j}, \\
&\xi_{k,j}^\star(\theta,\theta^\prime) = \xi_{k,j}(\theta,\theta^\prime) - \bm\Xi_{k,j}(\theta,\bm\theta_k^\star)\{\bm\Xi_{k,j}(\bm\theta_k^\star,\bm\theta_k^\star)+\mvec I_{n_k}\}^{-1} \bm\Xi_{k,j}(\bm\theta_k^\star,\theta^\prime), \\
& a_k = a_0 + n_k/2,\\
& b_{k,j} = b_0+\hat{\mvec X}_{k,j}^\intercal\{\bm\Xi_{k,j}(\bm\theta_k^\star,\bm\theta_k^\star)+\mvec I_{n_k}\}^{-1}\hat{\mvec X}_{k,j}/2,
\label{updates}
\end{align}
where $n_k=\sum_{i=1}^n\mathds 1_k\{z_i\}$, $\hat{\mvec X}_{k,j}\in\mathbb R^{n_k}$ is the subset of values of $\hat{\mvec X}_j$ for which $z_i=k$, and $\bm\theta_k^\star\in\mathbb R^{n_k}$ is the vector $\bm\theta$, restricted to the entries such that $z_i=k$. Furthermore, $\bm\Xi_{k,j}$ is a vector-valued and matrix-valued extension of $\xi_{k,j}$, such that $[\bm\Xi_{k,j}(\bm\theta,\bm\theta^\prime)]_{\ell,\ell^\prime}=\xi_{k,j}(\theta_\ell,\theta^\prime_{\ell^\prime})$. The structure of the GP-IG yields an analytic expression for the posterior predictive distribution for a new observation $\vec x^\ast=(x^\ast_1,\dots,x^\ast_d)$ in community $z^\ast$,
\begin{equation}
{\hat x}^\ast_j\vert z^\ast,\vec z,\bm\theta,\theta^\ast,\hat{\mvec X} \sim t_{2a_{z^\ast}}\left(\mu_{z^\ast,j}^{\star}(\theta^\ast),\frac{b_{z^\ast,j}}{a_{z^\ast}}\left\{1+\xi_{z^\ast,j}^{\star}(\theta^\ast, \theta^\ast)\right\}\right), \label{marginal_pred}
\end{equation}
where $t_\nu(\mu,\sigma)$ denotes a Student's $t$ distribution with $v$ degrees of freedom, mean $\mu$ and scale parameter $\sigma$.
Furthermore, the prior probabilities $\bm\eta$ for the community assignments can be integrated out, obtaining
\begin{equation}
p(\vec z) = \frac{\Gamma(\nu)\prod_{k=1}^K \Gamma(n_k+\nu/K)}{\Gamma(\nu/K)^K\Gamma(n+\nu)}, \label{marginal_z}
\end{equation}
where $n_k=\sum_{i=1}^n \mathds 1_k\{z_i\}$.
The two distributions \eqref{marginal_pred} and \eqref{marginal_z} are key components for the 
Bayesian inference algorithm discussed in the next section.

\subsection{Posterior inference} \label{sec:mcmc}

After marginalisation of the pairs $(f_{k,j},\sigma^2_{k,j})$ and $\bm\eta$, inference is limited to the community allocations $\vec z$ and latent parameters $\bm\theta$. 
The marginal posterior distribution $p(\vec z,\bm\theta\mid\hat{\mvec X})$ is analytically intractable; therefore, inference is performed using collapsed Metropolis-within-Gibbs Markov Chain Monte Carlo (MCMC) sampling. 
In this work, MCMC methods are used, but an alternative inferential algorithm often deployed in the community-detection literature is variational Bayesian inference \citep[see, for example,][]{Latouche12}, which is also applicable to 
GPs \citep{Cheng17}. 

For the community allocations $\vec z$, the Gibbs sampling step uses the following decomposition:
\begin{equation}
p(z_i=k\mid \vec z^{-i}, \hat{\mvec X},\bm\theta) \propto p(z_i=k\mid\vec z^{-i}) p(\hat{\vec x}_i\mid z_i=k,\vec z^{-i},\bm\theta,\hat{\mvec X}^{-i}),
\end{equation}
where the superscript $-i$ denotes that the $i$-th row (or element) is removed from the corresponding matrix (or vector). 
Using \eqref{marginal_z}, the first term is
\begin{equation}
p(z_i=k\mid\vec z^{-i})=\frac{n^{-i}_k+\nu/K}{n-1+\nu}.
\end{equation}
For the second term, using \eqref{marginal_pred}, the posterior predictive distribution for $\hat{\vec x}_i$ given $z_i=k$ can be written as the product of $d$ independent Student's $t$ distributions, where
\begin{equation}
\hat x_{i,j}\vert z_i=k,\vec z^{-i},\bm\theta,\hat{\mvec X}^{-i} \sim t_{2a_k^{-i}}\left(\mu_{k,j}^{\star-i}(\theta_i),\frac{b_{k,j}^{-i}}{a_k^{-i}}\left\{1+\xi_{k,j}^{\star-i}(\theta_i, \theta_i)\right\}\right). \label{t_post}
\end{equation}
Note that the quantities $\mu_{k,j}^{\star-i},\xi_{k,j}^{\star-i}, a_k^{-i}$ and $b_{k,j}^{-i}$ are calculated as described in \eqref{updates}, \textit{excluding} the contribution of the $i$-th node. 

In order to mitigate identifiability issues, it is necessary to assume that some of the parameters are known a priori. For example, assuming for each community $k$ that $f_{k,1}(\theta_i)=\theta_i$, corresponding to a linear model in $\theta$ with no intercept and unit slope in the first dimension, gives the predictive distribution: 
\begin{equation}
\hat x_{i,1}\vert z_i=k,\vec z^{-i},\bm\theta,\hat{\mvec X}^{-i} \sim t_{2a_{k,j}^{-i}}\left(\theta_i, 
\frac{1}{a_k^{-i}} \left\{ b_0 + \frac{1}{2}\sum_{h\neq i:z_h=k} (\hat x_{h,1} - \theta_h)^2 \right\} 
\right). \label{t_post2}
\end{equation}

Finally, for updates to $\mvec\theta_i$, a standard Metropolis-within-Gibbs step can be used. For a proposed value $\theta^\ast$ sampled from a proposal distribution $q(\cdot\mid\theta_i)$, the acceptance probability takes the value
\begin{equation}
\min\left\{1,\frac{p(\hat{\vec x}_i\vert z_i,\vec z^{-i},\theta^\ast,\bm\theta^{-i},\hat{\mvec X}^{-i})p(\theta^\ast)q(\theta_i\vert\theta^\ast)}{p(\hat{\vec x}_i\vert z_i,\vec z^{-i},\theta_i,\bm\theta^{-i},\hat{\mvec X}^{-i})p(\theta_i)q(\theta^\ast\vert\theta_i)}\right\}. \label{prop}
\end{equation} 
The proposal distribution $q(\theta^\ast\vert\theta_i)$ in this work is a normal distribution $\mathbb N(\theta^\ast\vert\theta_i,\sigma^2_\ast)$, $\sigma^2_\ast\in\mathbb R_+$, implying that the ratio of proposal distributions in \eqref{prop} cancels out by symmetry.

\subsection{Inference on the number of communities $K$}

So far, it has been assumed that the number of communities $K$ is known. The LSBM prior specification \eqref{prior_structure} naturally admits a prior distribution on the number of communities $K$. Following \cite{SannaPassino20_sbm}, it could be assumed:
\begin{equation}
K\sim\text{Geometric}(\omega),
\end{equation}
where $\omega\in(0,1)$. The MCMC algorithm is then augmented with two additional moves for posterior inference on $K$:
\begin{enumerate*}[label=(\roman*)]
\item split or merge two communities; and
\item add or remove an empty community.
\end{enumerate*}
An alternative approach when $K$ is unknown could also be a nonparametric overlapping mixture of Gaussian processes \citep{Ross13}. 
For simplicity, in the next two sections, it will be initially assumed that \emph{all} communities have the same functional form, corresponding, for example, to the same basis functions $\bm\phi_{k,j}(\cdot)$ for dot product kernels \eqref{dot_kernel}. Then, in Section~\ref{sec:prior_kernel}, the algorithm will be extended to admit a prior distribution on the community-specific kernels. 

\subsubsection{Split or merge two communities} \label{split_merge}

In this case, the proposal distribution follows \cite{SannaPassino20_sbm}.
First, two nodes $i$ and $j$ are sampled randomly. For simplicity, assume $z_i\leq z_j$. If $z_i\neq z_j$, then the two corresponding communities are merged into a unique cluster: all nodes in community $z_j$ are assigned to $z_i$. Otherwise, if $z_i=z_j$, the cluster is split into two different communities, proposed as follows: 
\begin{enumerate*}[label=(\roman*)]
\item node $i$ is assigned to community $z_i$ ($z_i^\ast=z_i$), and node $j$ to community $K^\ast=K+1$ ($z_j^\ast=K^\ast$); 
\item the remaining nodes in community $z_i$ are allocated in random order to clusters $z_i^\ast$ or $z_j^\ast$ according to their posterior predictive distribution \eqref{t_post} or \eqref{t_post2}, restricted to the two communities, and calculated sequentially.
\end{enumerate*}
It follows that the proposal distribution $q(K^\ast,\vec z^\ast\vert K, \vec z)$ for a split move corresponds to the product of renormalised posterior predictive distributions, leading to the following acceptance probability:
\begin{equation}
\alpha(K^\ast,\vec z^\ast\vert K,\vec z) = \min\left\{ 1, \frac{p(\hat{\mvec X}\vert K^\star,\vec z^\star,\bm\theta) p(\vec z^\ast\vert K^\ast)p(K^\ast)}
{p(\hat{\mvec X}\vert K,\vec z,\bm\theta)p(\vec z\vert K)p(K) q(K^\ast,\vec z^\ast\vert K, \vec z)}
\right\}, \label{acc_split}
\end{equation}
where $p(\hat{\mvec X}\vert K,\vec z,\bm\theta)$ is the marginal likelihood. Note that the ratio of marginal likelihoods only depends on the two communities involved in the split and merge moves. Under \eqref{eq:xhat} and \eqref{model_gp}, the community-specific marginal on the $j$-th dimension is:
\begin{equation}
\hat{\mvec X}_{k,j} \vert K, \vec z, \bm\theta \sim t_{2a}\left(\vec 0_{n_k}, \frac{b_0}{a_0}\{\bm\Xi_{k,j}(\bm\theta_k^\star, \bm\theta_k^\star) + \mvec I_{n_k}\} \right), \label{marginal_posterior}
\end{equation}
where the notation is identical to \eqref{updates}, and the Student’s $t$ distribution is $n_k$-dimensional, with mean equal to the identically zero vector $\vec 0_\cdot$. The full marginal $p(\hat{\mvec X}\vert K,\vec z,\bm\theta)$ is the product of marginals \eqref{marginal_posterior} on all dimensions and communities. 
If $f_{k,1}(\theta_i)=\theta_i$, \textit{cf.} \eqref{t_post2}, the marginal likelihood is
\begin{equation}
p(\hat{\mvec X}_{k,1}\vert K,\vec z,\bm\theta) = \frac{\Gamma(a_{n_k})}{\Gamma(a_0)} \frac{b_0^{a_0}}{b_{n_k}^{a_{n_k}}} (2\pi)^{-n_k/2},
\end{equation}
where $a_{n_k} = a_0+n_k/2$, $b_{n_k} = b_0 + \sum_{i:z_i=k} (\hat x_{i,1} - \theta_i)^2/2$.  

\subsubsection{Add or remove an empty community} \label{sec:empty}

When adding or removing an empty community, the acceptance probability is:
\begin{equation}
  \alpha(K^\ast\vert K )= \min\left\{ 1, \frac{p(\vec z\vert K^\ast)p(K^\ast)q_\varnothing}{p(\vec z\vert K)p(K)}
  \right\},
  \label{eq:empty}
\end{equation}
where $q_\varnothing=q(K\vert K^\ast,\vec z)/q(K^\ast\vert K,\vec z)$ is the proposal ratio, equal to 
\begin{enumerate*}[label=(\roman*)]
\item $q_\varnothing=2$ if the proposed number of clusters $K^\ast$ equals the number of non-empty communities in $\vec z$;
\item $q_\varnothing=0.5$ if there are no empty clusters in $\vec z$; and
\item $q_\varnothing=1$ otherwise.
\end{enumerate*}
Note that the acceptance probability is identical to \citet[Section~4.3]{SannaPassino20_sbm}, and it does not depend on the marginal likelihoods. 

\subsection{Inference with different community-specific kernels} \label{sec:prior_kernel}

When communities are assumed to have \emph{different} functional forms, it is required to introduce a prior distribution $p(\bm\xi_k),\ \bm\xi_k=(\xi_{k,1},\dots,\xi_{k,d})$, on the GP kernels, supported on one or more classes of possible kernels $\mathcal K$. Under this formulation, a proposal to change the community-specific kernel could be introduced. 
Conditional on the allocations $\vec z$, the $k$-th community is assigned a kernel $\bm\xi^\ast=(\xi^\ast_1,\dots,\xi^\ast_d)$ with probability:
\begin{equation}
p(\bm\xi_k=\bm\xi^\ast\vert\hat{\mvec X},K,\vec z,\bm\theta) \propto p(\bm\xi^\ast)\prod_{j=1}^d p(\hat{\mvec X}_{k,j}\vert K,\vec z,\bm\theta, \xi^\ast_j),
\end{equation}
normalised for $\bm\xi^\ast\in\mathcal K$, where $p(\hat{\mvec X}_{k,j}\vert K,\vec z,\bm\theta, \xi^\ast_j)$ is the marginal \eqref{marginal_posterior} calculated under kernel $\xi^\ast_j$.
The prior distribution $p(\bm\xi^\ast)$ could also be used as proposal for the kernel of an empty community (\textit{cf.} Section~\ref{sec:empty}). Similarly, in the merge move (\textit{cf.} Section~\ref{split_merge}), the GP kernel could be sampled at random from the two kernels assigned to $z_i$ and $z_j$, correcting the acceptance probability \eqref{acc_split} accordingly. 

\section{Applications and results} \label{sec:results}

Inference for the LSBM is tested on synthetic LSBM data and on three real world networks. As discussed in Section \ref{sec:mcmc}, the first dimension $\hat{\mvec X}_1$ is assumed to be linear in $\theta_i$, with no intercept and unit slope. It follows that $\theta_i$ is initialised to $\hat{x}_{i,1}+\varepsilon_i$, where $\varepsilon_i\sim\mathbb N(0,\sigma^2_\varepsilon)$, for a small $\sigma^2_\varepsilon$, usually equal to $0.01$. Note that such an assumption links the proposed Bayesian model for LSBM embeddings to Bayesian errors-in-variables models \citep{Dellaportas95}. In the examples in this section, the kernel function is assumed to be of the dot product form \eqref{dot_kernel}, with Zellner's $g$-prior such that $\bm\Delta_{k,j} = n^2\{\bm\Phi_{k,j}(\bm\theta)^\intercal\bm\Phi_{k,j}(\bm\theta)\}^{-1}$, where $\bm\Phi_{k,j}(\bm\theta)\in\mathbb R^{n\times q_{k,j}}$ such that the $i$-th row corresponds to $\bm\phi_{k,j}(\theta_i)$. For the remaining parameters: $a_0=1$, $b_0=0.001$, $\mu_\theta=\sum_{i=1}^n \hat x_{i,1}/n$, $\sigma^2_\theta=10$, $\nu=1$, $\omega=0.1$.

The community allocations are initialised using $k$-means with $K$ groups, unless otherwise specified. 
The final cluster configuration is estimated from the output of the MCMC algorithm described in Section~\ref{sec:mcmc}, using the estimated posterior similarity between nodes $i$ and $j$, $\hat\pi_{ij}=\hat{\mathbb P}(z_i=z_j\mid\hat{\mvec X})=\sum_{s=1}^{M} \mathds 1_{z^\star_{i,s}}\{z^\star_{j,s}\}/M$, where $M$ is the total number of posterior samples and $z^\star_{i,s}$ is the $s$-th sample for $z_i$. The posterior similarity is not affected by the issue of label switching \citep{Jasra05}. The clusters are subsequently estimated using hierarchical clustering with average linkage, with distance measure $1-\hat\pi_{ij}$ \citep{medve}. The quality of the estimated clustering compared to the true partition, when available, is evaluated using the Adjusted Rand Index \citep[ARI,][]{Hubert85}. The results presented in this section are based on $M=\numprint{10000}$ posterior samples, with $\numprint{1000}$ burn-in period.

\subsection{Simulated data: stochastic blockmodels} \label{sec:sbms}

First, the performance of the LSBM and related inferential procedures is assessed on simulated data from two common models for graph clustering: stochastic blockmodels and degree-corrected stochastic blockmodels (\textit{cf.} Examples~\ref{sbm_sim} and \ref{dcsbm_sim}). Furthermore, a graph is also simulated from a blockmodel with quadratic latent structure (\textit{cf.} Example~\ref{quad_sim}). In particular, it is evaluated whether the Bayesian inference algorithm for LSBMs recovers the latent communities from the two-dimensional embeddings plotted in Figure~\ref{sim_graphs} (Section~\ref{sec:lsbm}), postulating the correct latent functions corresponding to the three generative models under consideration. 
The Gaussian process kernels implied by these three models are 
(i) $\xi_{k,j}(\theta,\theta^\prime)=\Delta_{k,j}\in\mathbb R_+$, $k\in\{1,2\},\ j\in\{1,2\}$ for the SBM; 
(ii) $\xi_{k,2}(\theta,\theta^\prime)=\theta\theta^\prime\Delta_{k,2},\ k\in\{1,2\}$ for the DCSBM;
(iii) $\xi_{k,2}(\theta,\theta^\prime)=(\theta,\theta^2)\bm\Delta_{k,2}(\theta^\prime,\theta^{\prime 2})^\intercal$ 
for the quadratic LSBM.
For the DCSBM and quadratic LSBM, it is also assumed that $\hat{\mvec X}_1$ is linear in $\theta_i$ with no intercept and unit slope, corresponding to $f_{k,1}(\theta)=\theta$. 

The performance of the inferential algorithm is also compared to alternative methods for spectral graph clustering: Gaussian mixture models (GMM; which underly the assumption of a SBM generative model) on the ASE $\hat{\mvec X}$, GMMs on the row-normalised ASE $\tilde{\mvec X}$ \citep[see, for example,][]{RubinDelanchy17}, and GMMs on a spherical coordinate transformation of the embedding \citep[SCSC, see][]{SannaPassino20}. Note that the two latter methods postulate a DCSBM generative model. The ARI is averaged across 1000 initialisations of the GMM inferential algorithm. Furthermore, the proposed methodology is compared to a kernel-based clustering technique using Gaussian processes: the parsimonious Gaussian process\footnote{PGP-EM was coded in \textit{python} starting from the PGP-DA implementation in the \textit{GitHub} repository \href{https://github.com/mfauvel/PGPDA}{\texttt{mfauvel/PGPDA}}.} model (PGP) of \cite{Bouveyron15}, fitted using the EM algorithm (PGP-EM). In all the implementations of PGP-EM in Section~\ref{sec:results}, the RBF kernel is used, with variance parameter chosen via grid-search, with the objective to maximise the resulting ARI. The responsibilities in the EM algorithm are initialised using the predictive probabilities obtained from a Gaussian mixture model fit on the ASE or row-normalised ASE, chosen according to the largest ARI. Also, the LSBM is compared to the hierarchical Louvain (HLouvain in Table~\ref{clust_table_droso}) algorithm\footnote{Implemented in \textit{python} in the library \textit{scikit-network}.} for graphs, corresponding to hierarchical clustering by successive instances of the Louvain algorithm \citep{Blondel08}. Finally, the LSBM is also compared to hierarchical clustering with complete linkage and Euclidean distance (HClust in Table~\ref{clust_table_droso}), applied on $\hat{\mvec X}$. 

\begin{table*}[t]
\centering
\scalebox{0.85}{
\begin{tabular}{c | cccccccc}
\toprule
Method & LSBM$(\hat{\mvec X})$ & GMM$(\hat{\mvec X})$ & GMM$(\tilde{\mvec X})$ & SCSC$(\hat{\mvec X})$ & PGP$(\hat{\mvec X})$ & HLouvain & HClust$(\hat{\mvec X})$ \\
\midrule
SBM & 1.0 & 1.0 & 1.0 & 1.0 & 1.0 & 1.0 & 1.0 \\
DCSBM & 0.853 & 0.802 & 0.838 & 0.887 & 0.842 & 0.827 & 0.411 \\
Quadratic LSBM & 0.838 & 0.620 & 0.712 & 0.636 & 0.691 & 0.582 & 0.101 \\
\bottomrule
\end{tabular}
}
\caption{ARI for communities estimated using LSBM and alternative methodologies on the embeddings in Figure~\ref{sim_graphs}.}
\label{table_sim_ari}
\end{table*}

The results are presented in Table~\ref{table_sim_ari}. The performance of the LSBM appears to be on par with alternative methodologies for SBMs and DCSBMs. This is expected, since the LSBM does not have competitive advantage over alternative methodologies for inference under such models. In particular, inferring LSBMs with constant latent functions using the algorithm in Section~\ref{sec:mcmc} is equivalent to Bayesian inference in Gaussian mixture models with normal-inverse gamma priors. For DCSBMs, the LSBM is only marginally outperformed by spectral clustering on spherical coordinates \citep[SCSC,][]{SannaPassino20}.
On the other hand, LSBMs largely outperform competing methodologies when quadratic latent structure is observed in the embedding.
It must be remarked that the coefficients of the latent functions used to generate the quadratic latent structure model in Figure~\ref{quad_plot} were chosen to approximately reproduce the curves in the cyber-security example in Figure~\ref{maths_cive}. Despite the apparent simplicity of the quadratic LSBM used in this simulation, its practical relevance is evident from plots of real-world network embeddings, and it is therefore important to devise methodologies to estimate its communities correctly.
Furthermore, LSBMs could also be used when the number of communities $K$ is unknown. In the examples in this section, inference on $K$ using the MCMC algorithm overwhelmingly suggests $K=2$, the correct value used in the simulation. 

\subsection{Simulated data: Hardy-Weinberg LSBM} \label{sec:sim}

\begin{figure*}[t]
\centering
\begin{subfigure}[t]{0.325\textwidth}
\centering
\caption{}
\includegraphics[width=\textwidth]{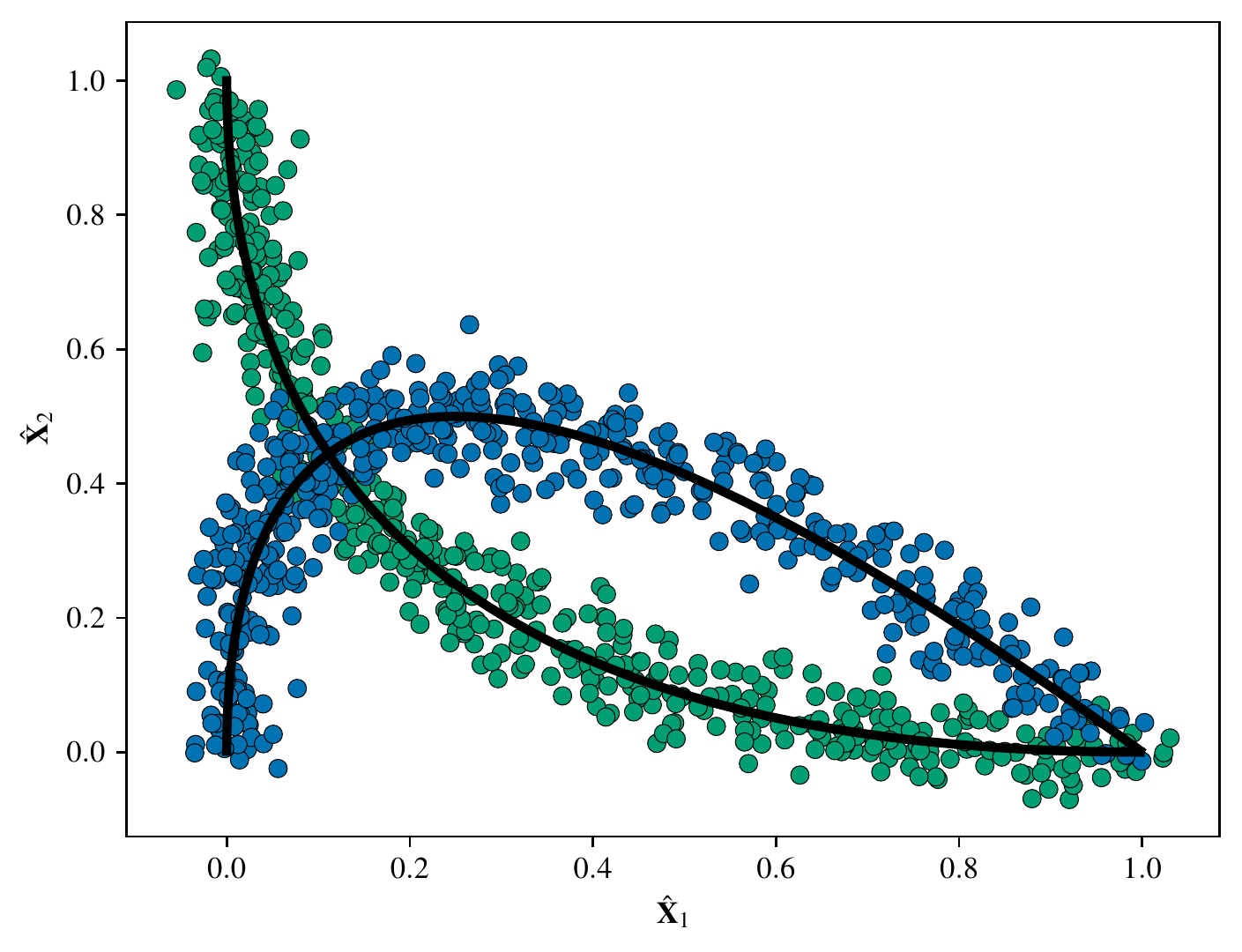}
\end{subfigure}
\begin{subfigure}[t]{0.325\textwidth}
\centering
\caption{}
\includegraphics[width=\textwidth]{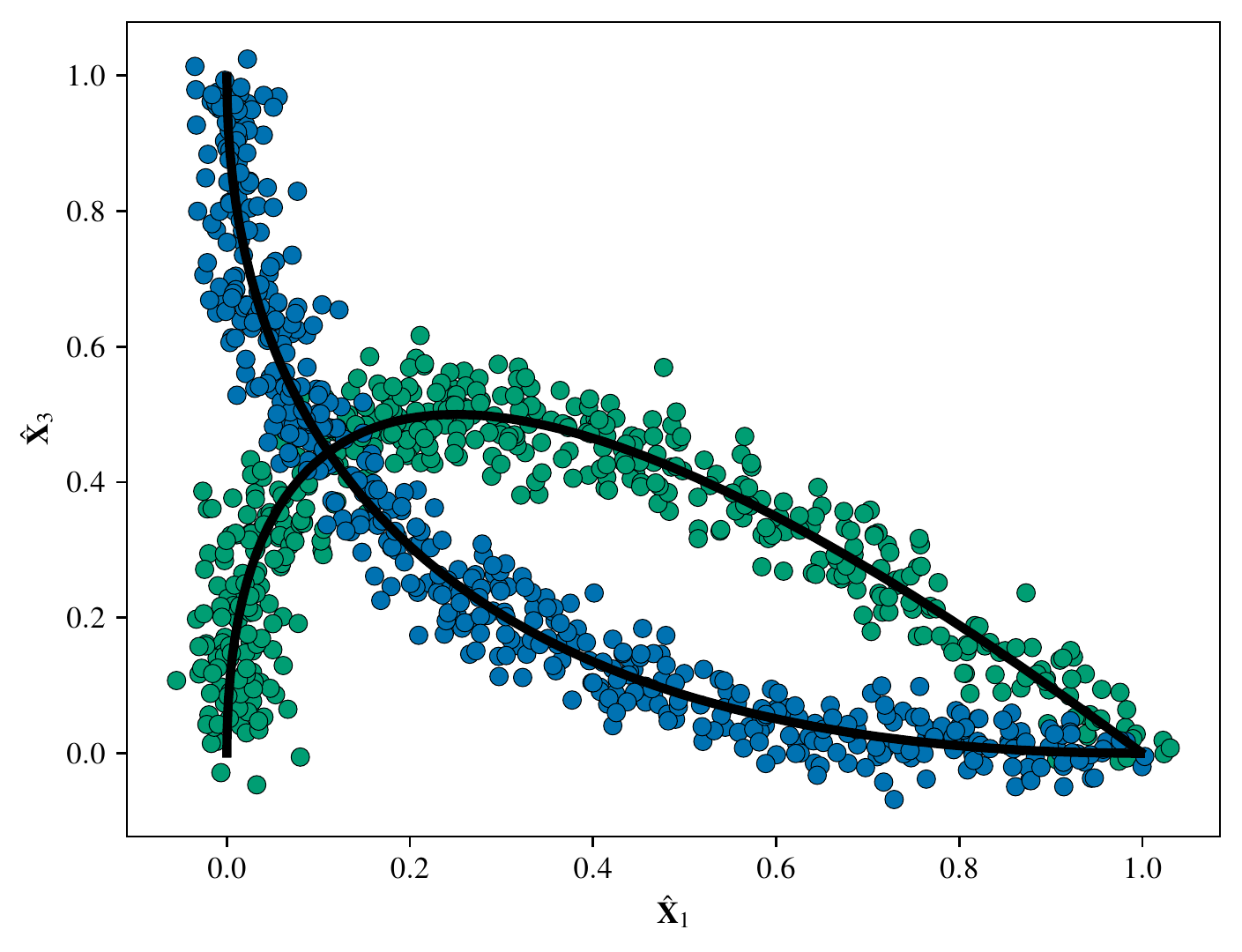}
\end{subfigure}
\begin{subfigure}[t]{0.325\textwidth}
\centering
\caption{}
\includegraphics[width=\textwidth]{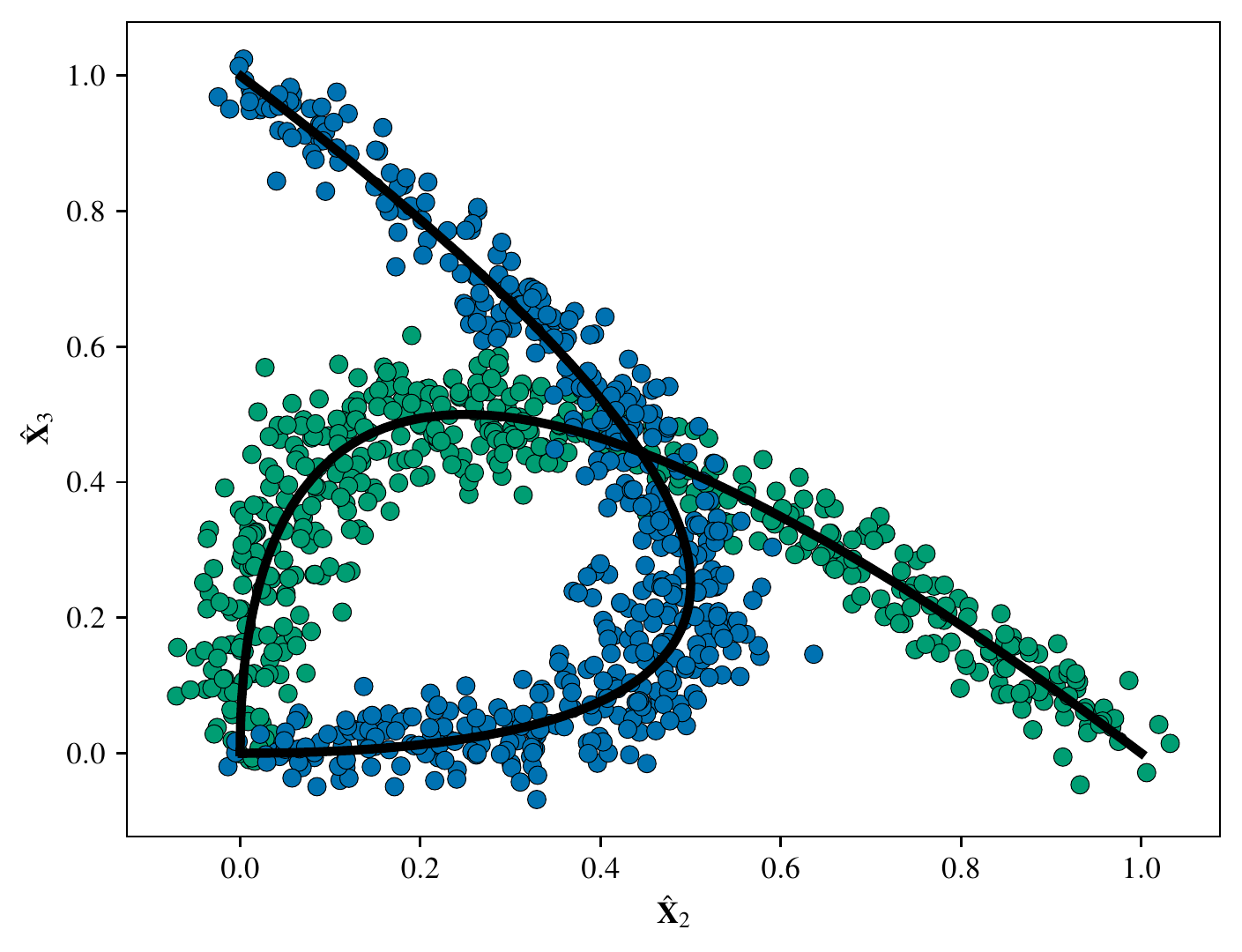}
\end{subfigure}
\caption{Scatterplots of $\{\hat{\mvec X}_1,\hat{\mvec X}_2,\hat{\mvec X}_3\}$, coloured by community, and true underlying latent positions (in black).}
\label{sim_scatter}
\end{figure*}

Second, the performance of the LSBM is assessed on simulated data on a Hardy-Weinberg LSBM. In \cite{Athreya21} and \cite{Trosset20}, a Hardy-Weinberg latent structure model is used, with $\mathcal G=[0,1]$ and $\vec f(\theta)=\{\theta^2,2\theta(1-\theta),(1-\theta)^2\}$. A permutation of the Hardy-Weinberg curve is considered here for introducing community structure. A graph with $n=\numprint{1000}$ nodes is simulated, with $K=2$ communities of equal size. Each node is assigned a latent score $\theta_i\sim\text{Uniform}(0,1)$, which is used to obtain the latent position $\vec x_i$ through the mapping $\vec f_{z_i}(\theta_i)$. In particular:
\begin{align}
\vec f_1(\theta) =\{(1-\theta)^2,\theta^2,2\theta(1-\theta)\}, \\
\vec f_2(\theta)=\{\theta^2,2\theta(1-\theta),(1-\theta)^2\}.
\end{align}
Using the latent positions, the graph adjacency matrix is then generated under the random dot product graph kernel $\mathbb P(A_{ij}=1 \mid \vec x_i,\vec x_j)=\vec x_i^\intercal\vec x_j$. The resulting scatterplot of the latent positions estimated via ASE is plotted in Figure~\ref{sim_scatter}. For visualisation, the estimated latent positions $\hat{\mvec X}$ have been aligned to the true underlying latent positions $\mvec X$ using a Procrustes transformation \citep[see, for example,][]{Mardia16}. 

\begin{figure*}[t]
\centering
\begin{subfigure}[t]{0.49\textwidth}
\centering
\caption{Quadratic latent functions}
\includegraphics[width=\textwidth]{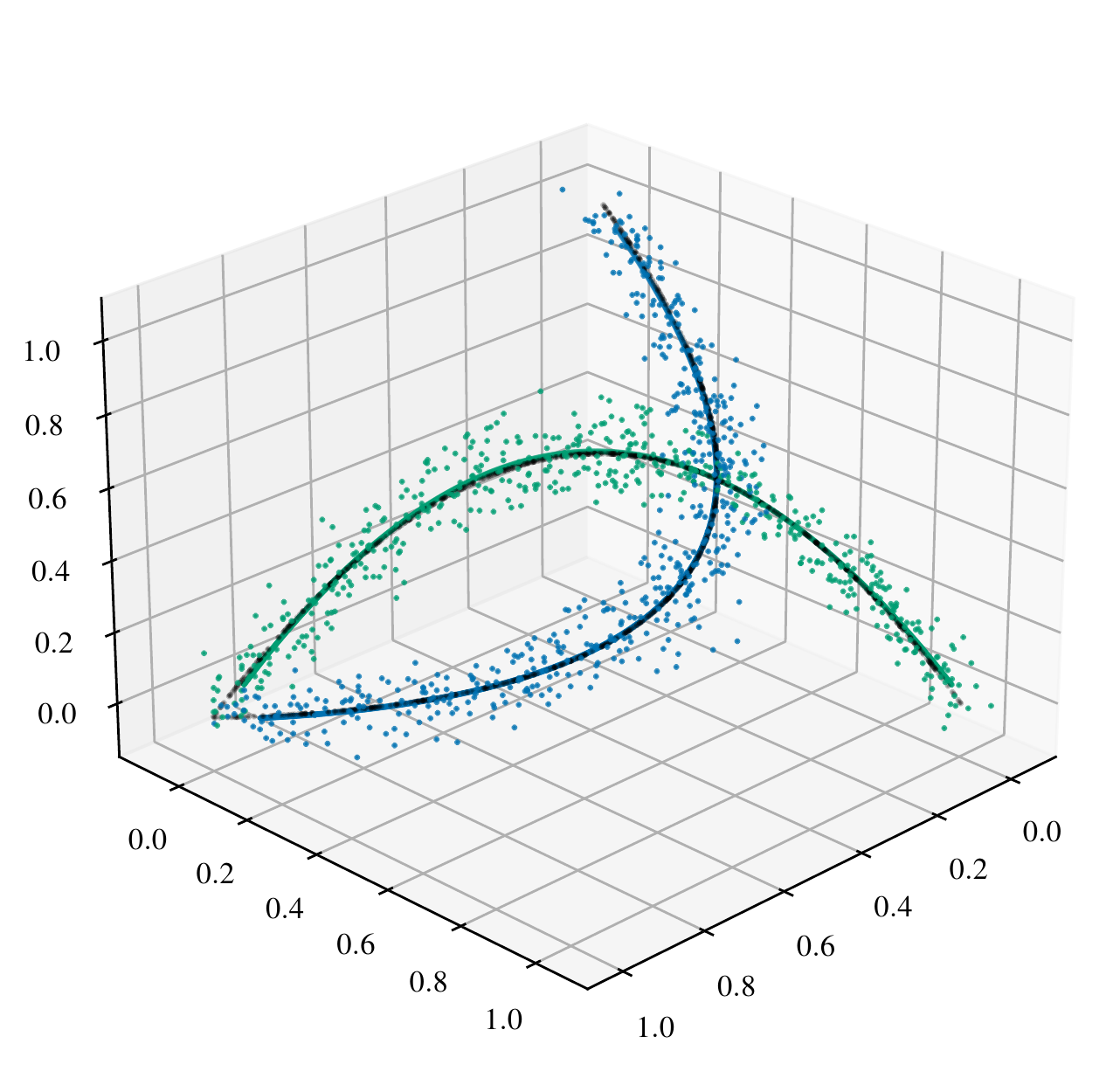}
\label{3d_poly}
\end{subfigure}
\begin{subfigure}[t]{0.49\textwidth}
\centering
\caption{Cubic latent functions, linearity in first dimension}
\includegraphics[width=\textwidth]{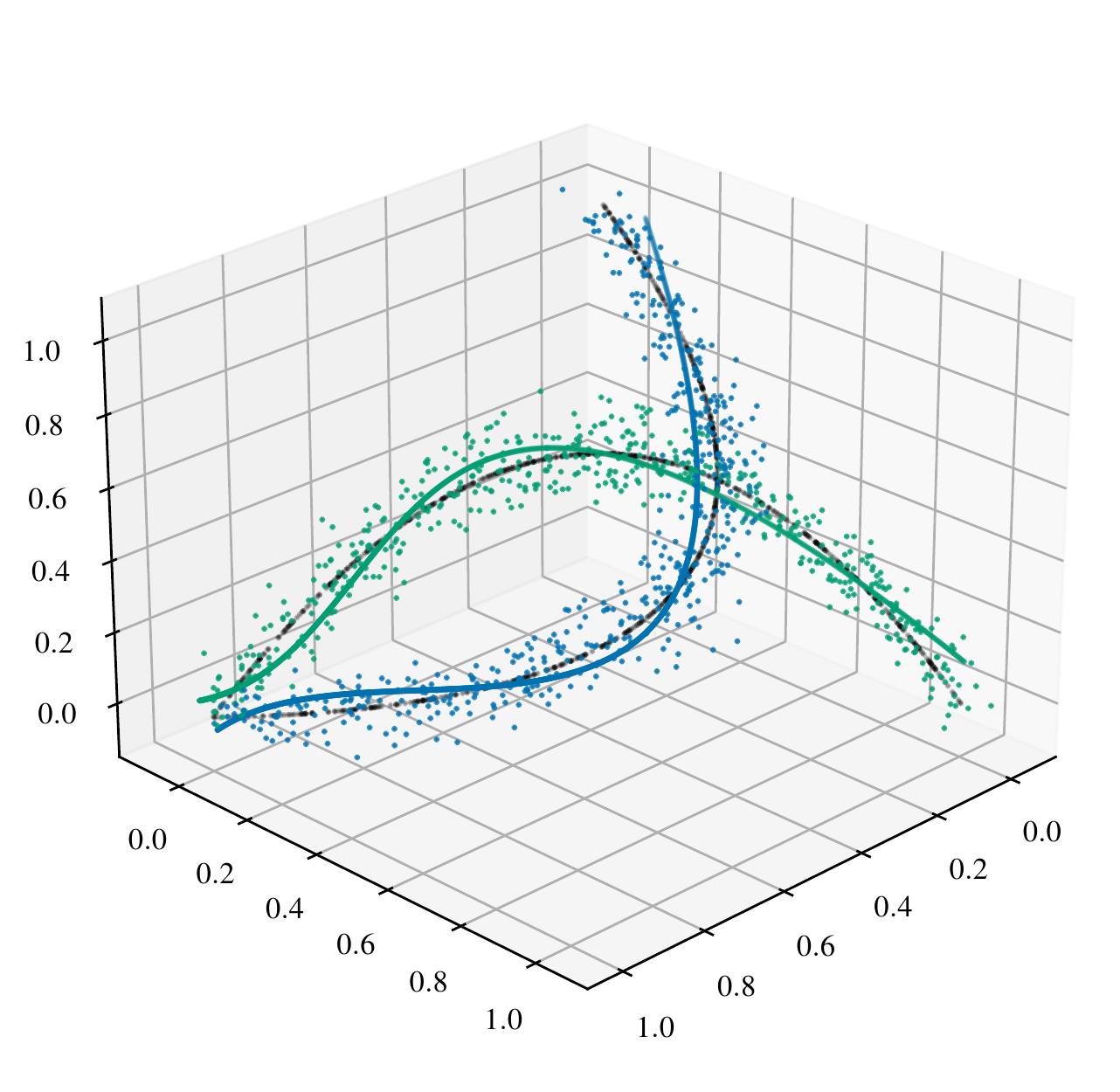}
\label{3d_cubic}
\end{subfigure}
\caption{Scatterplot of $\hat{\mvec X}$ and estimated generating curves obtained from the estimated clustering, coloured by community, and true underlying latent positions (in {black}).}
\label{scatter_sim}
\end{figure*}

The inferential procedure is first run assuming that the parametric form of the underlying latent function is fully known. Therefore, the kernels are set to $\xi_{k,j}(\theta,\theta^\prime)=(1,\theta,\theta^2)\bm\Delta_{k,j}(1,\theta^\prime,\theta^{\prime 2})^\intercal,\ k=1,2,\ j=1,2,3$. Figure~\ref{3d_poly} shows the best-fitting curves for the two estimated communities after MCMC, which are almost indistinguishable from the true underlying latent curves. The Markov Chain was initialised setting $\theta_i=\abs{\hat x_{i,1}}^{1/2}$, and obtaining initial values of the allocations $\vec z$ from $k$-means. The resulting ARI is $0.7918$, which is not perfect since some of the nodes at the intersection between the two curves are not classified correctly, but corresponds to approximately $95\%$ of nodes correctly classified. The number of communities is estimated to be $K=2$ from the inferential algorithm, the correct value used in the simulation.  

If the underlying functional relationship is unknown, a realistic guess could be given by examining the scatterplots of the embedding. The scatterplots in Figure~\ref{sim_scatter} show that, assuming linearity in $\theta_i$ on $\hat{\mvec X}_1$ with no intercept and unit slope, a quadratic or cubic polynomial function is required to model $\hat{\mvec X}_2$ and $\hat{\mvec X}_3$. Therefore, for the purposes of the MCMC inference algorithm in Section~\ref{sec:mcmc}, $f_2(\cdot)$ and $f_3(\cdot)$ are assumed to be cubic functions, corresponding to the Gaussian process kernel $\xi_{k,j}(\theta,\theta^\prime)=(1,\theta,\theta^2,\theta^3)\bm\Delta_{k,j}(1,\theta^\prime,\theta^{\prime2},\theta^{\prime3})^\intercal$, $j\in\{2,3\}$, whereas $f_{k,1}(\theta) = \theta,\ k=1,\dots,K$. The curves corresponding to the estimated clustering 
are plotted in Figure~\ref{3d_cubic}.
Also in this case, the algorithm is able to approximately recover the curves that generated the graph, and the number of communities $K=2$. The imperfect choice of the latent functions makes the ARI decrease to $0.6687$, which still corresponds to over $90\%$ of nodes correctly classified. 

Table~\ref{hw_alternative} presents further comparisons with the alternative techniques discussed in the previous section, showing that only LSBMs recover a good portion of the communities. This is hardly surprising, since alternative methodologies 
do not take into account the quadratic latent functions generating the latent positions, and therefore fail to capture the underlying community structure.  

\begin{table*}[t]
\centering
\scalebox{0.85}{
\begin{tabular}{c | cccccccc}
\toprule
Method & LSBM$(\hat{\mvec X})$ & LSBM$(\hat{\mvec X})$ & GMM$(\hat{\mvec X})$ & GMM$(\tilde{\mvec X})$ & SCSC$(\hat{\mvec X})$ & PGP$(\hat{\mvec X})$ & HLouvain & HClust$(\hat{\mvec X})$\\
& Quadratic & Cubic + Linear \\
\midrule
ARI & 0.791 & 0.668 & -0.001 & -0.001 & 0.000 & 0.063 & -0.001 & 0.053 \\
\bottomrule
\end{tabular}
}
\caption{ARI for communities estimated using LSBM and alternative methodologies on the embeddings in Figure~\ref{sim_scatter}.}
\label{hw_alternative} 
\end{table*}

\subsection{Undirected graphs: Harry Potter enmity graph}

\begin{figure}[b]
\centering
\includegraphics[width=0.475\textwidth]{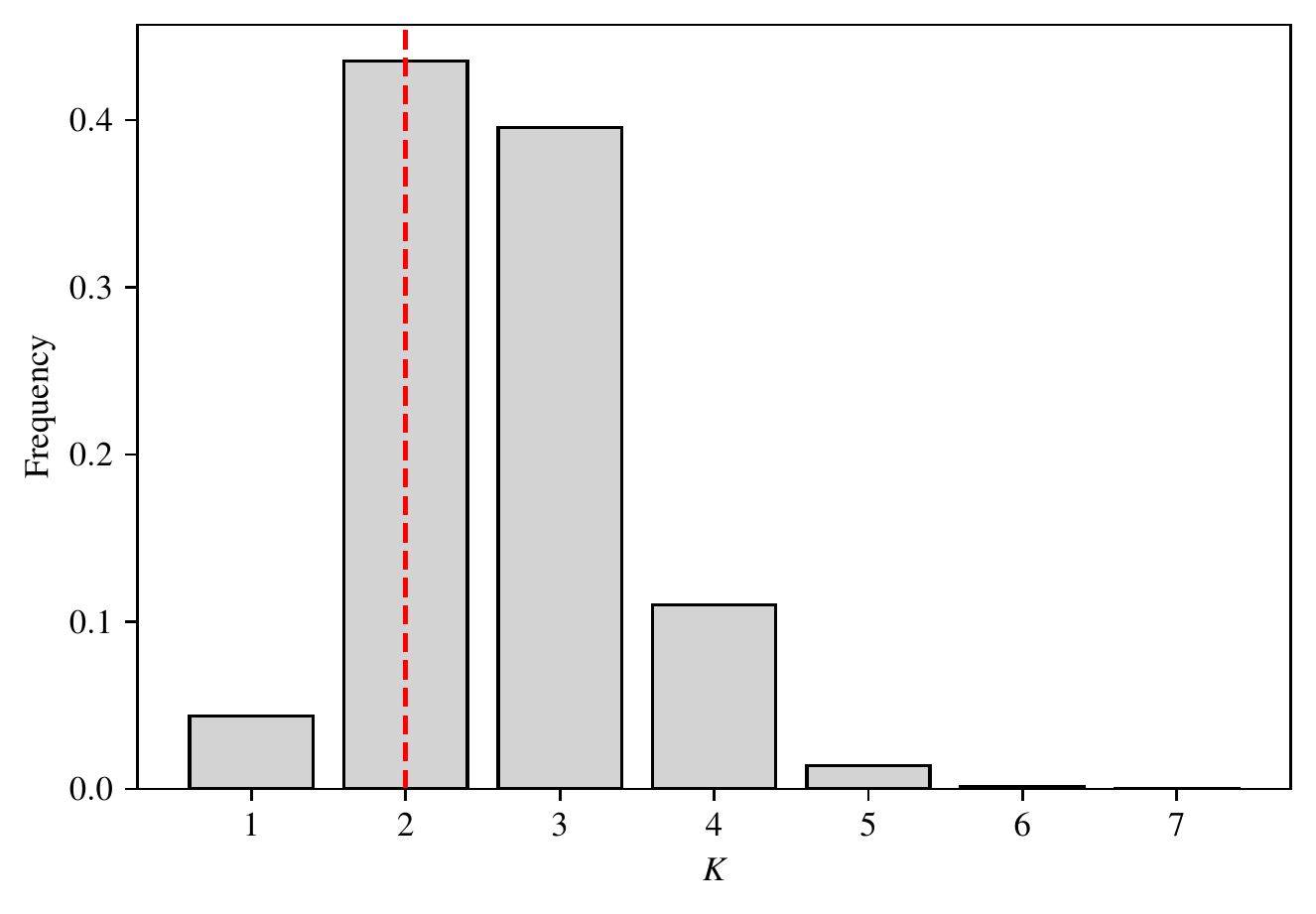}
\caption{Estimated marginal posterior for the number of non-empty clusters $K$ on the Harry Potter network, under linear kernels.}
\label{harry_Khist}
\end{figure}

\begin{figure*}[t]
\centering
\begin{subfigure}[t]{0.49\textwidth}
\centering
\caption{Linear latent functions}
\includegraphics[width=\textwidth]{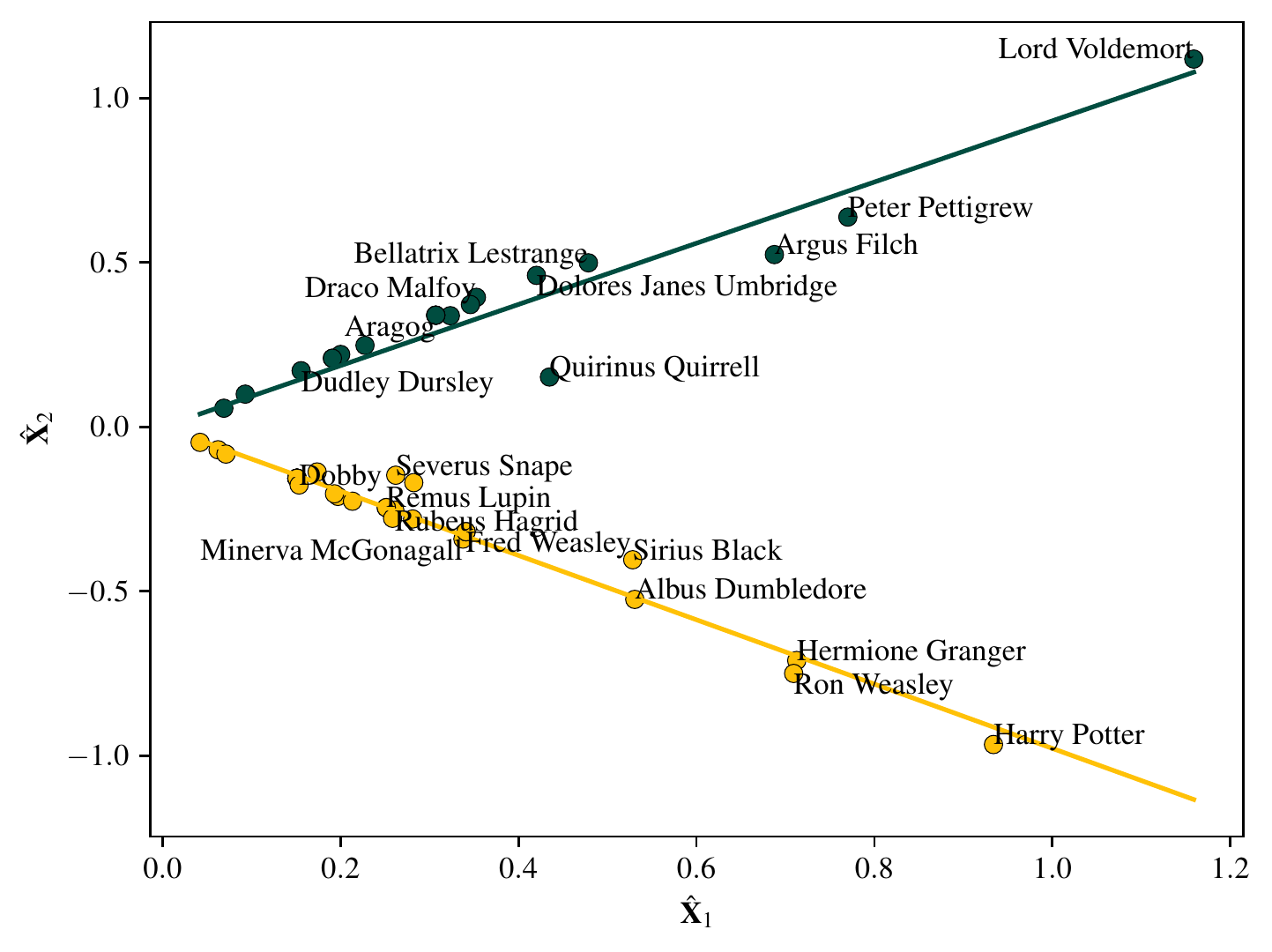}
\label{harry_linear}
\end{subfigure}
\begin{subfigure}[t]{0.49\textwidth}
\centering
\caption{Splines with cubic truncated power basis}
\includegraphics[width=\textwidth]{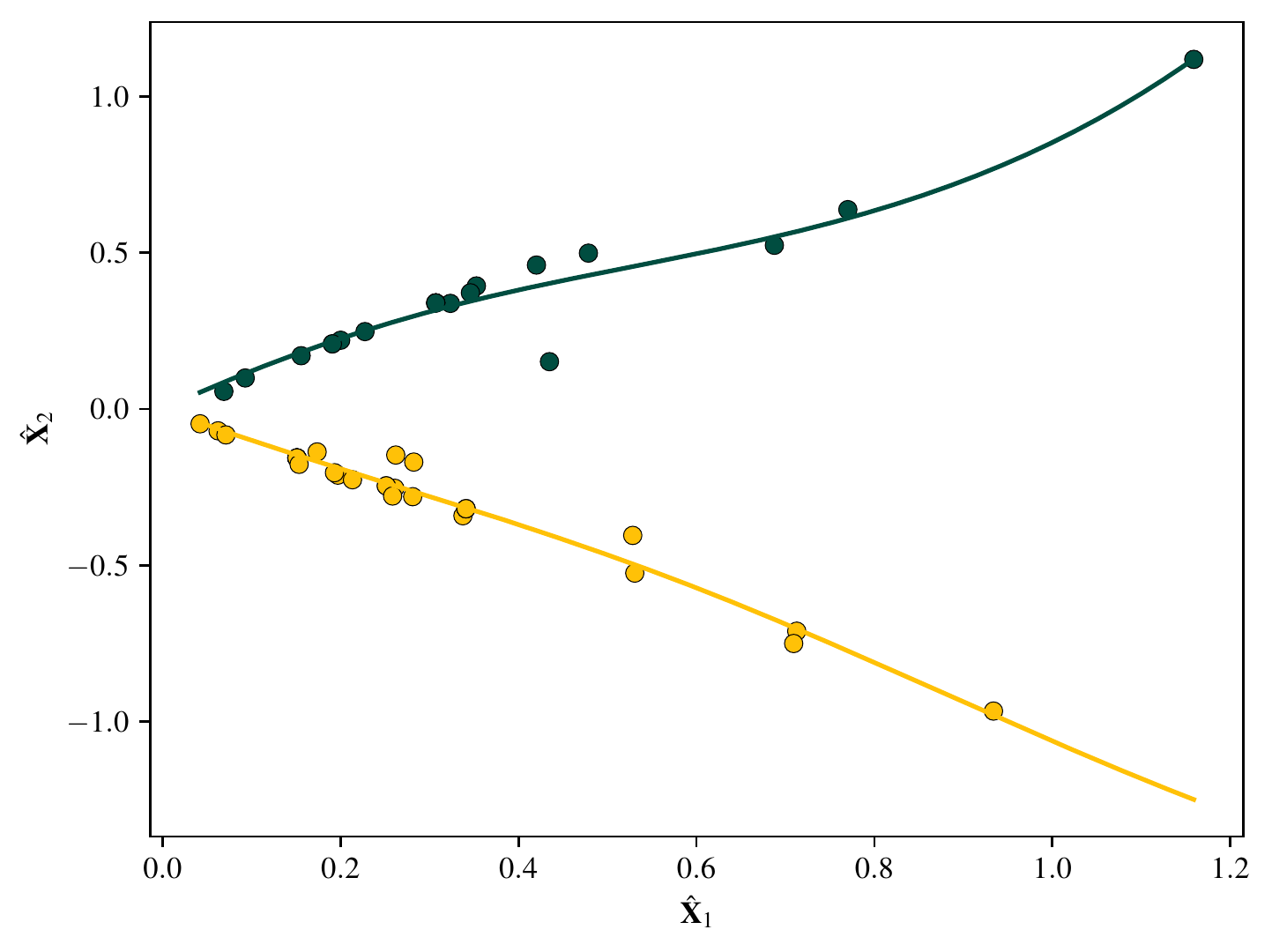}
\label{harry_splines}
\end{subfigure}
\caption{Scatterplots of $\{\hat{\mvec X}_2,\hat{\mvec X}_3\}$ vs. $\hat{\mvec X}_1$, coloured by community allocation, and best fitting line and cubic truncated power spline passing through the origin, obtained from the estimated clustering.}
\label{scatter_harry}
\end{figure*}

The LSBM is also applied to the Harry Potter enmity graph\footnote{The network is publicly available in the \textit{GitHub} repository \href{https://github.com/efekarakus/potter-network}{\texttt{efekarakus/potter-network}}.}, an undirected network with $n=51$ nodes representing characters of J.K. Rowling's series of fantasy novels. In the graph, $A_{ij}=A_{ji}=1$ if the characters $i$ and $j$ are enemies, and $0$ otherwise. A degree-corrected stochastic blockmodel represents a reasonable assumption for such a graph \citep[see, for example,][]{Modell21}: Harry Potter, the main character, and Lord Voldemort, his antagonist, attract many enemies, resulting in a large degree, whereas their followers are expected to have lower degree. The graph might be expected to contain $K=2$ communities, since Harry Potter's friends are usually Lord Voldemort's enemies in the novels.
Theorem~\ref{clt} suggests that 
the embeddings for a DCSBM are expected to appear as \textit{rays} passing through the origin. Therefore, $\vec f_k(\cdot)$ is assumed to be composed of linear functions such that $f_{k,1}(\theta)=\theta$ and $\xi_{k,2}(\theta,\theta^\prime)=\theta\theta^\prime\Delta_{k,2}$. The estimated posterior distribution of the number of non-empty clusters is plotted in Figure~\ref{harry_Khist}. It appears that $K=2$ is the most appropriate number of clusters. The resulting $2$-dimensional ASE embedding of $\mvec A$ and the estimated clustering obtained using the linear LSBM are pictured in Figure~\ref{harry_linear}. The two estimated communities roughly correspond to the Houses of Gryffindor and Slytherin. The inferential algorithm often admits $K=3$, with a singleton cluster for Quirinus Quirrell, which appears to be between the two main linear groups (\textit{cf.} Figure~\ref{harry_linear}). 

Alternatively, if a polynomial form for $\vec f_k(\cdot)$ is unknown, a more flexible model is represented by regression splines, which can be also expressed in the form \eqref{dot_kernel}. A common choice for $\bm\phi_{k,j}(\cdot)$ is a cubic truncated power basis 
$\bm\phi_{k,j}=(\phi_{k,j,1},\dots,\phi_{k,j,6})$, $\ell\in\mathbb Z_+$, such that:
\begin{align}
&\phi_{k,j,1}(\theta)=\theta,\ \phi_{k,j,2}(\theta)=\theta^2,\ \phi_{k,j,3}(\theta)=\theta^3, \\  &\phi_{k,j,3+\ell}(\theta)=(\theta-\tau_\ell)_+^3,\ \ell=1,2,3,
\label{spline_functions}
\end{align}
where $(\tau_1,\tau_2,\tau_3)$ are knots, and $(\cdot)_+=\max\{0,\cdot\}$. 
In this application, the knots were selected as three equispaced points in the range of $\hat{\mvec X}_1$.
The results are plotted in Figure~\ref{harry_splines}. 
Using either functional form, the algorithm is clearly able to recover the two communities, meaningfully clustering Harry Potter's and Lord Voldemort's followers. 

\subsection{Directed graphs: \textit{Drosophila} connectome} \label{droso_section}

LSBM are also useful to cluster the larval \textit{Drosophila} mushroom body connectome \citep{Eichler17}, a directed graph representing connections between $n=213$ neurons in the brain of a species of fly\footnote{The data are publicly available in the \textit{GitHub} repository \href{https://github.com/youngser/mbstructure}{\texttt{youngser/mbstructure}}.}. 
The right hemisphere mushroom body connectome contains of $K=4$ groups of neurons: Kenyon Cells, Input Neurons, Output Neurons and Projection Neurons. If two neurons are connected, then $A_{ij}=1$, otherwise $A_{ij}=0$, forming an asymmetric adjacency matrix $\mvec A\in\{0,1\}^{n\times n}$.
The network has been extensively analysed in \cite{Priebe17} and \cite{Athreya18}.

Following \cite{Priebe17} and \cite{Athreya18}, after applying the DASE (Definition~\ref{dase}) for $d=3$, a joint concatenated embedding $\hat{\mvec Y}=[\hat{\mvec X},\hat{\mvec X}^\prime]\in\mathbb R^{n\times 2d}$ is obtained from $\mvec A$.
Based on the analysis of \cite{Priebe17}, it should be assumed that three of the communities (Input Neurons, Output Neurons and Projection Neurons) correspond to a stochastic blockmodel, resulting in Gaussian clusters, whereas the Kenyon Cells form a quadratic curve with respect to the first dimension in the embedding space. Therefore, the kernel functions implied in \cite{Priebe17} are 
$\xi_{1,j}(\theta,\theta^\prime)=(\theta,\theta^2)\bm\Delta_{1,j}(\theta^\prime,\theta^{\prime2})^\intercal,\ j=2,\dots,2d$, with $f_{1,1}(\theta)=\theta$, for the first community (corresponding to the Kenyon Cells), and $\xi_{k,j}(\theta,\theta^\prime)=\Delta_{k,j},\ k=2,3,4,\ j=1,\dots,2d$ for the three remaining groups of neurons. 

\begin{figure*}[t]
\centering
\begin{subfigure}[t]{0.49\textwidth}
\centering
\caption{$\hat{\mvec Y}_2$ vs. $\hat{\mvec Y}_1$}
\includegraphics[width=\textwidth]{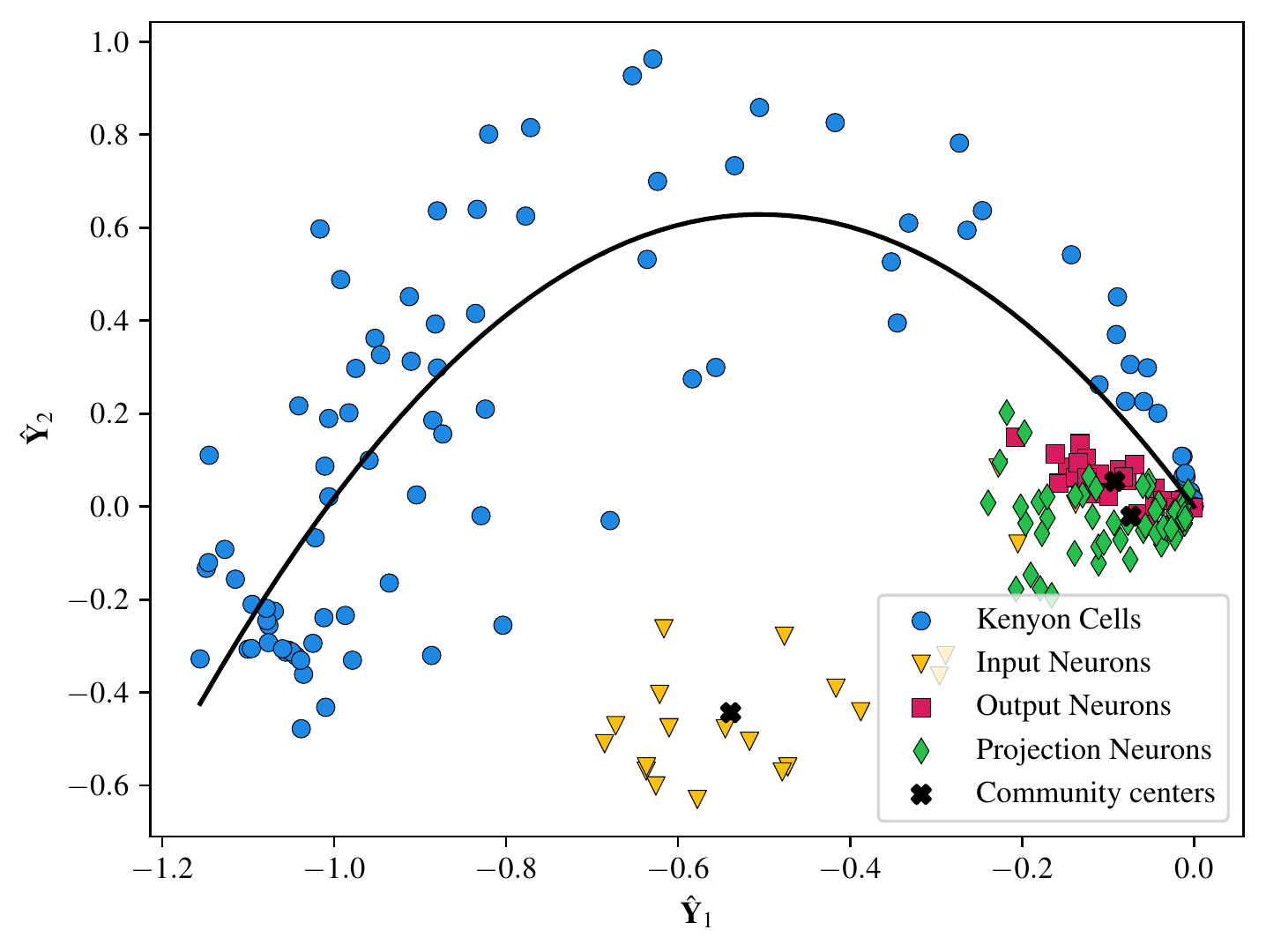}
\end{subfigure}
\begin{subfigure}[t]{0.49\textwidth}
\centering
\caption{$\hat{\mvec Y}_3$ vs. $\hat{\mvec Y}_1$}
\includegraphics[width=\textwidth]{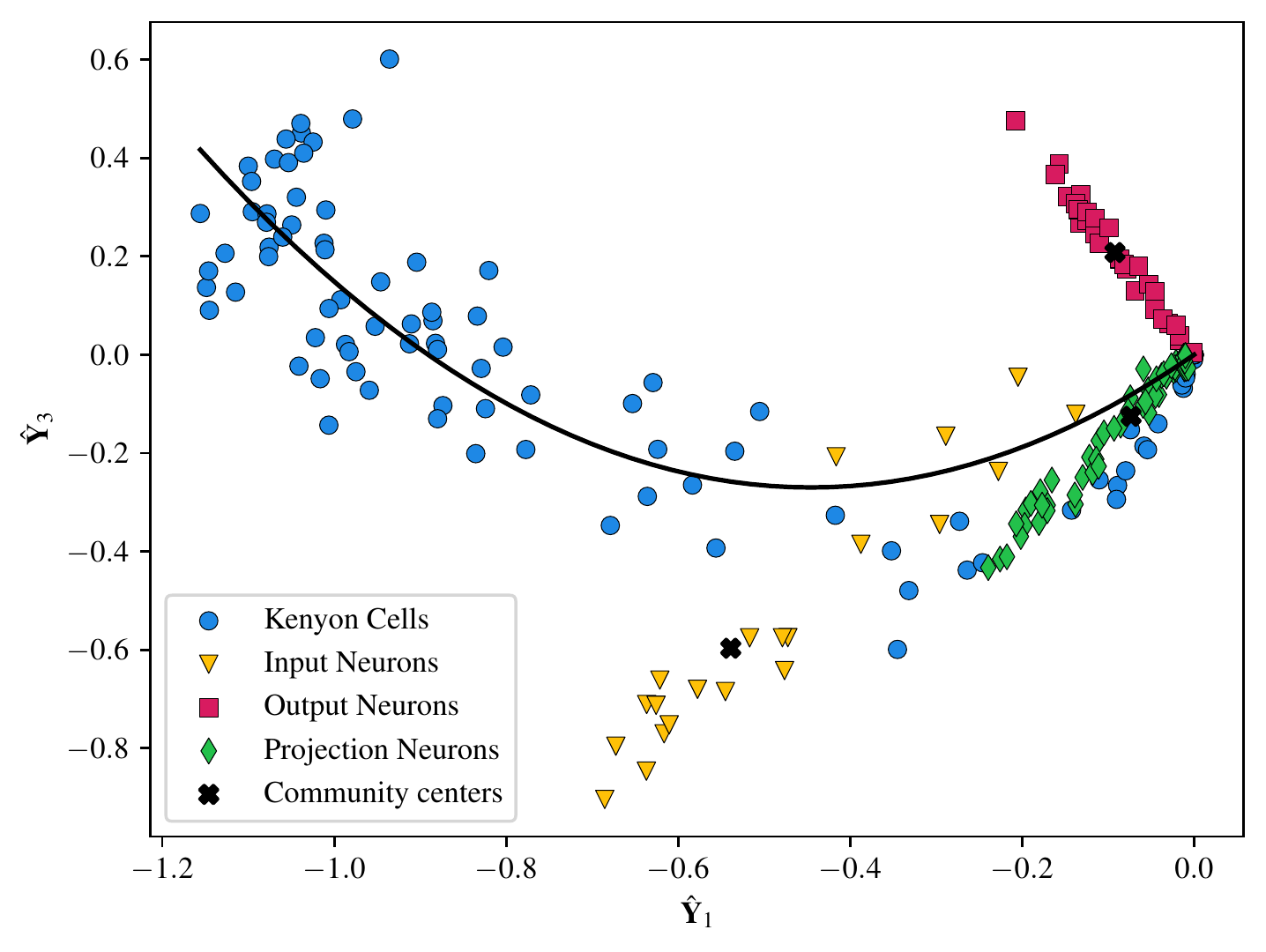}
\end{subfigure}
\centering
\begin{subfigure}[t]{0.325\textwidth}
\centering
\caption{$\hat{\mvec Y}_4$ vs. $\hat{\mvec Y}_1$}
\includegraphics[width=\textwidth]{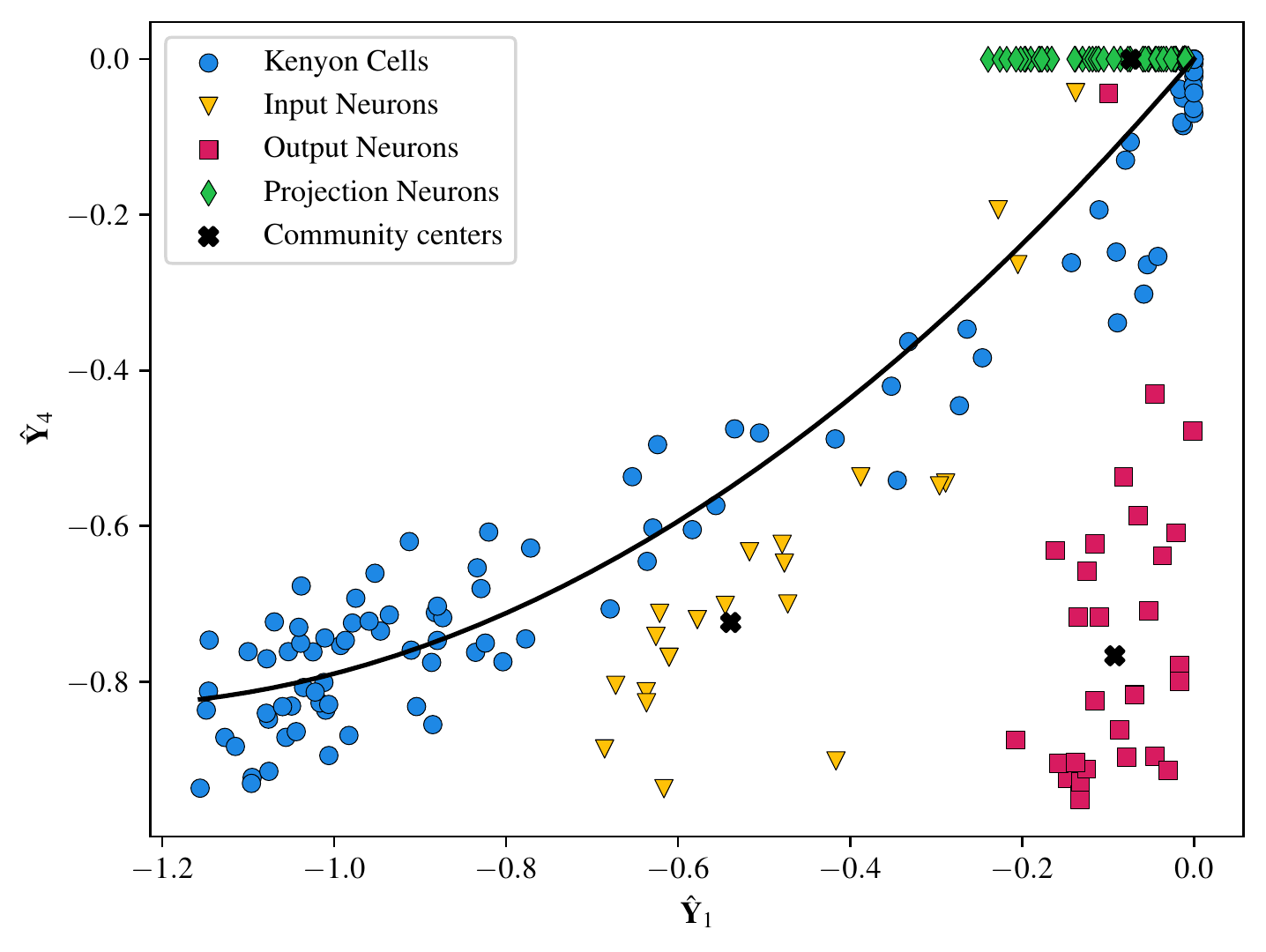}
\end{subfigure}
\centering
\begin{subfigure}[t]{0.325\textwidth}
\centering
\caption{$\hat{\mvec Y}_5$ vs. $\hat{\mvec Y}_1$}
\includegraphics[width=\textwidth]{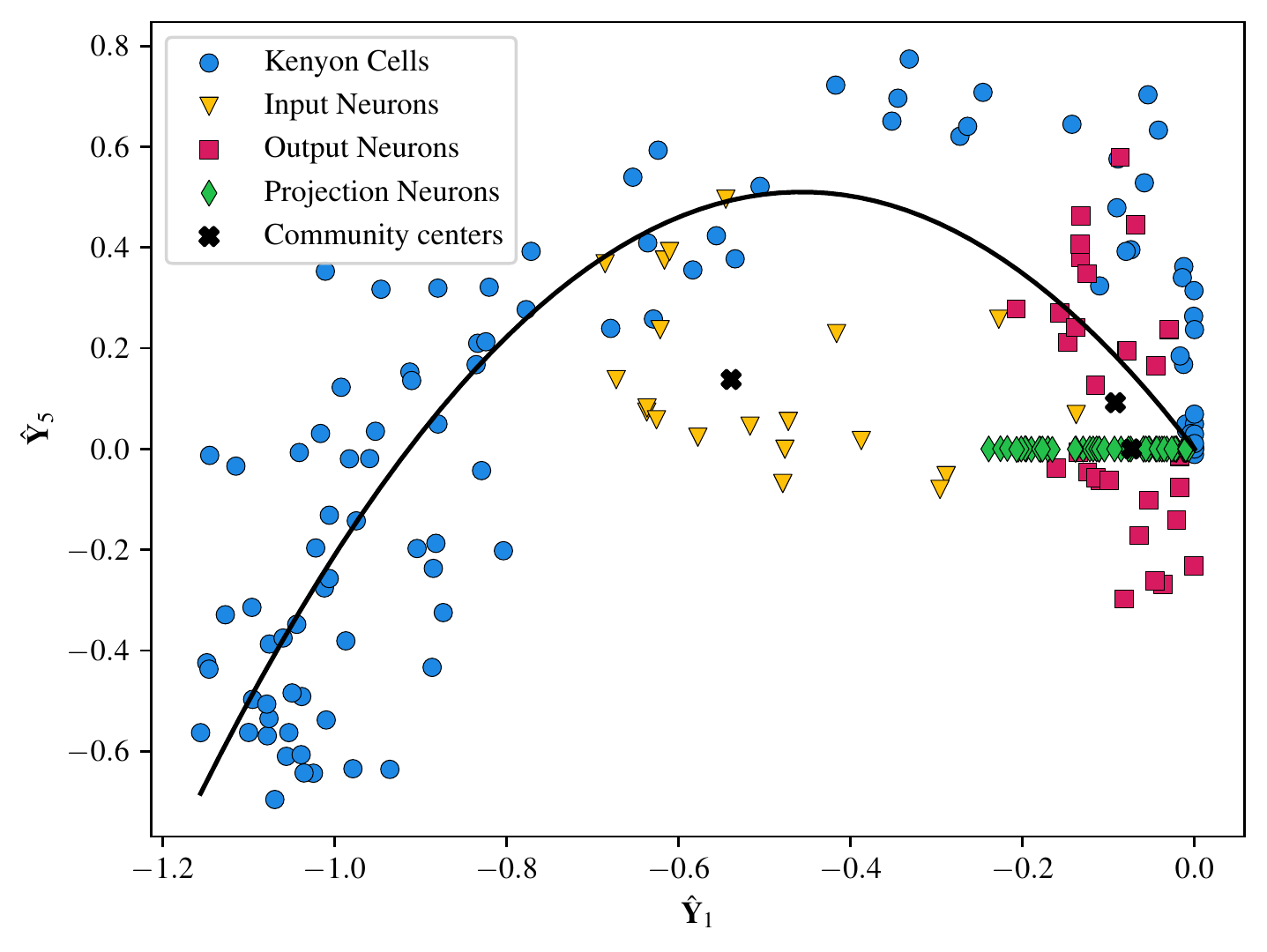}
\end{subfigure}
\begin{subfigure}[t]{0.325\textwidth}
\centering
\caption{$\hat{\mvec Y}_6$ vs. $\hat{\mvec Y}_1$}
\includegraphics[width=\textwidth]{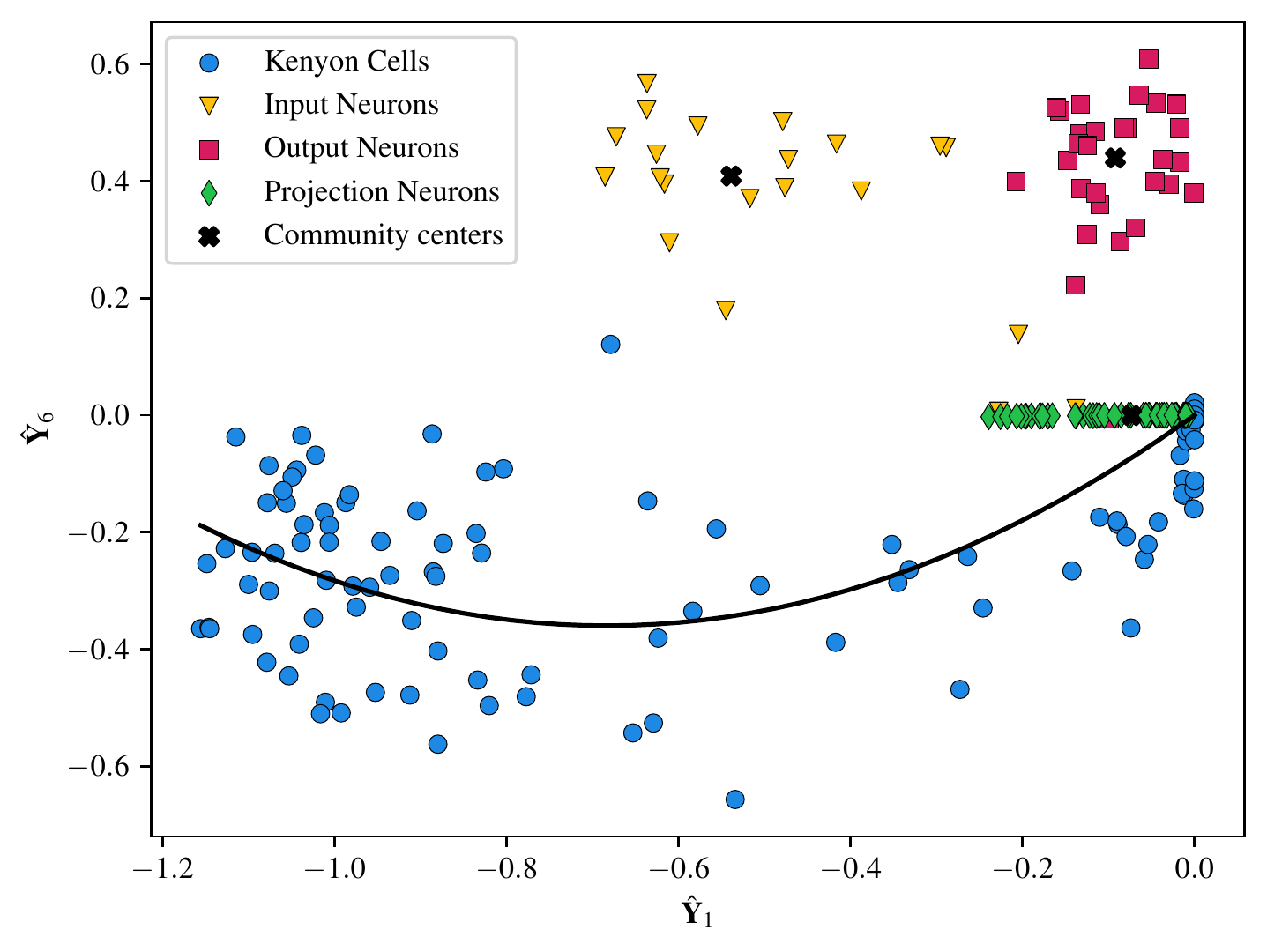}
\end{subfigure}
\caption{Scatterplots of $\{\hat{\mvec Y}_2,\hat{\mvec Y}_3,\hat{\mvec Y}_4,\hat{\mvec Y}_5,\hat{\mvec Y}_6\}$ vs. $\hat{\mvec Y}_1$, coloured by neuron type, and best fitting latent functions obtained from the estimated clustering.}
\label{scatters_droso}
\end{figure*}

Following the discussion in \cite{Priebe17}, the LSBM inferential procedure is initialised using $k$-means with $K=6$, and grouping three of the clusters to obtain $K=4$ initial groups. Note that, since the output of $k$-means is invariant to permutations of the $K=4$ labels for the community allocations, careful relabelling of the initial values is necessary to ensure that the Kenyon Cells effectively correspond to the first community which assumes a quadratic functional form. The most appropriate relabelling mapping is chosen here by repeatedly initialising the model with \textit{all} possible permutations of the labels, and choosing the permutation that maximises the marginal likelihood. Note that the marginal likelihood under the Bayesian model \eqref{model_gp} for LSBMs is analytically available in closed form \citep[see, for example,][]{Rasmussen06}. The results obtained after MCMC sampling are plotted in Figure~\ref{scatters_droso}. The estimated clustering has ARI $0.8643$, corresponding to only 10 misclassified nodes out of 213. 

\begin{figure*}[t]
\centering
\begin{subfigure}[t]{0.49\textwidth}
\centering
\caption{$\hat{\mvec Y}_2$ vs. $\hat{\mvec Y}_1$}
\includegraphics[width=\textwidth]{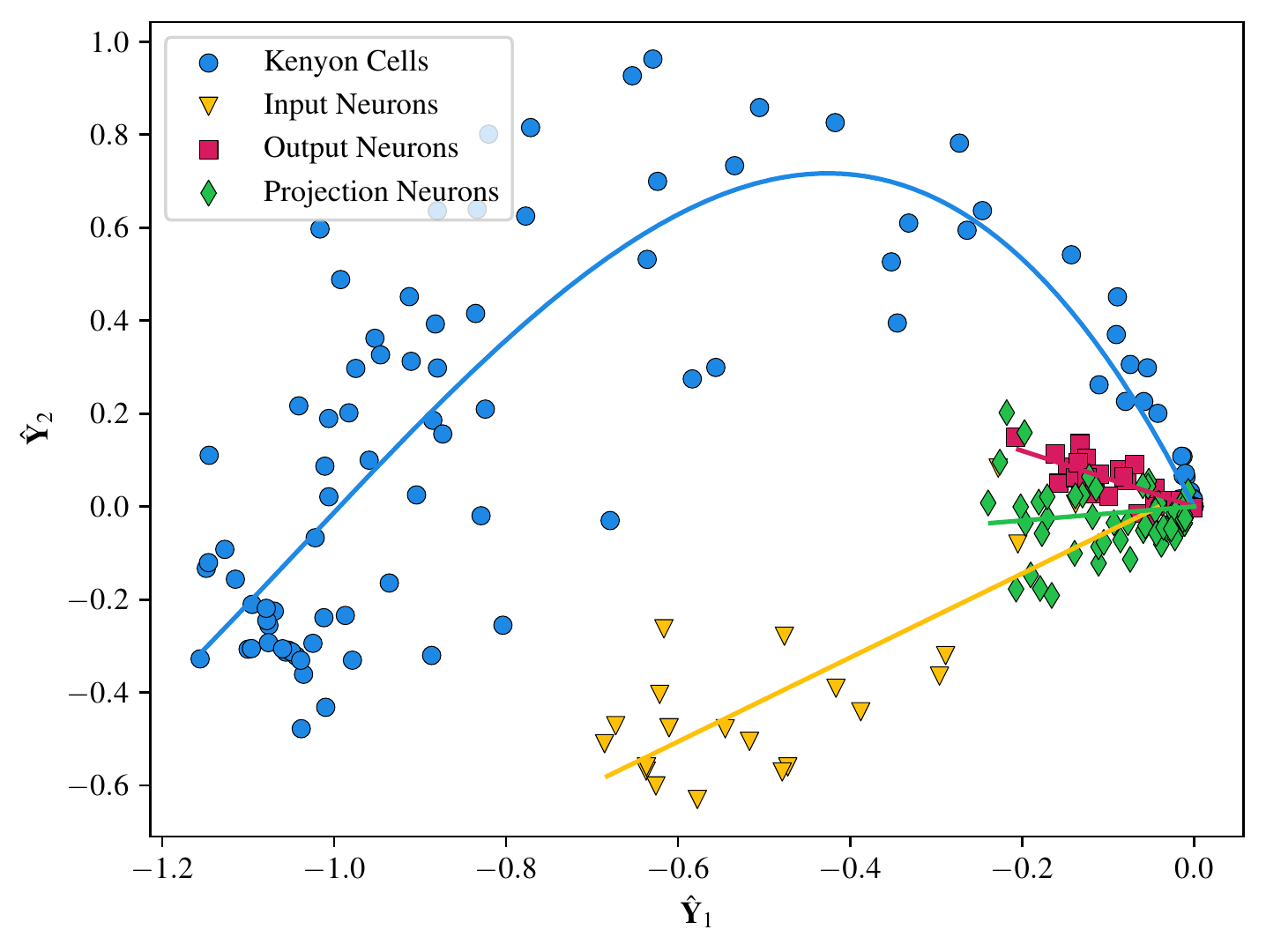}
\end{subfigure}
\begin{subfigure}[t]{0.49\textwidth}
\centering
\caption{$\hat{\mvec Y}_3$ vs. $\hat{\mvec Y}_1$}
\includegraphics[width=\textwidth]{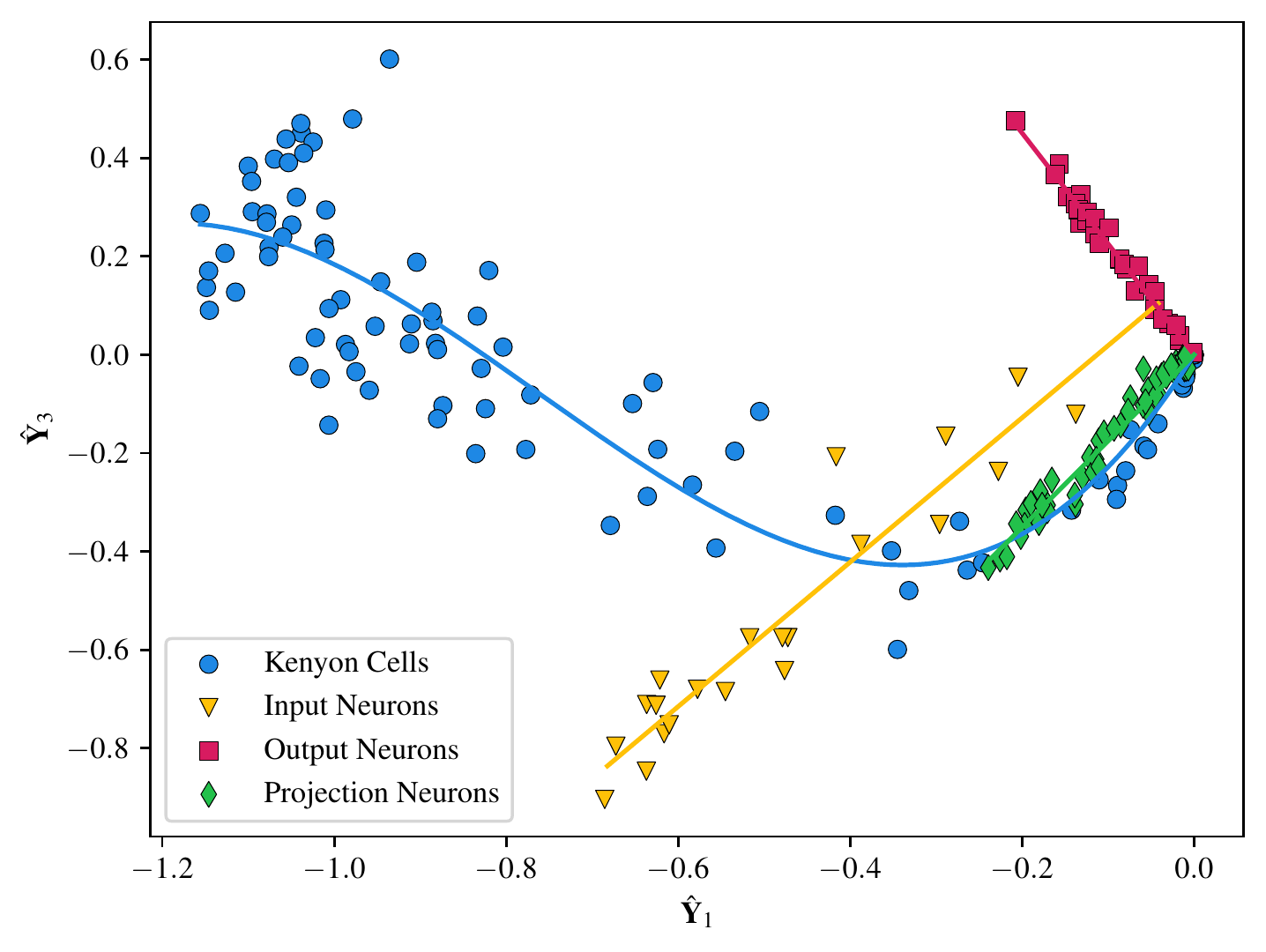}
\end{subfigure}
\centering
\begin{subfigure}[t]{0.325\textwidth}
\centering
\caption{$\hat{\mvec Y}_4$ vs. $\hat{\mvec Y}_1$}
\includegraphics[width=\textwidth]{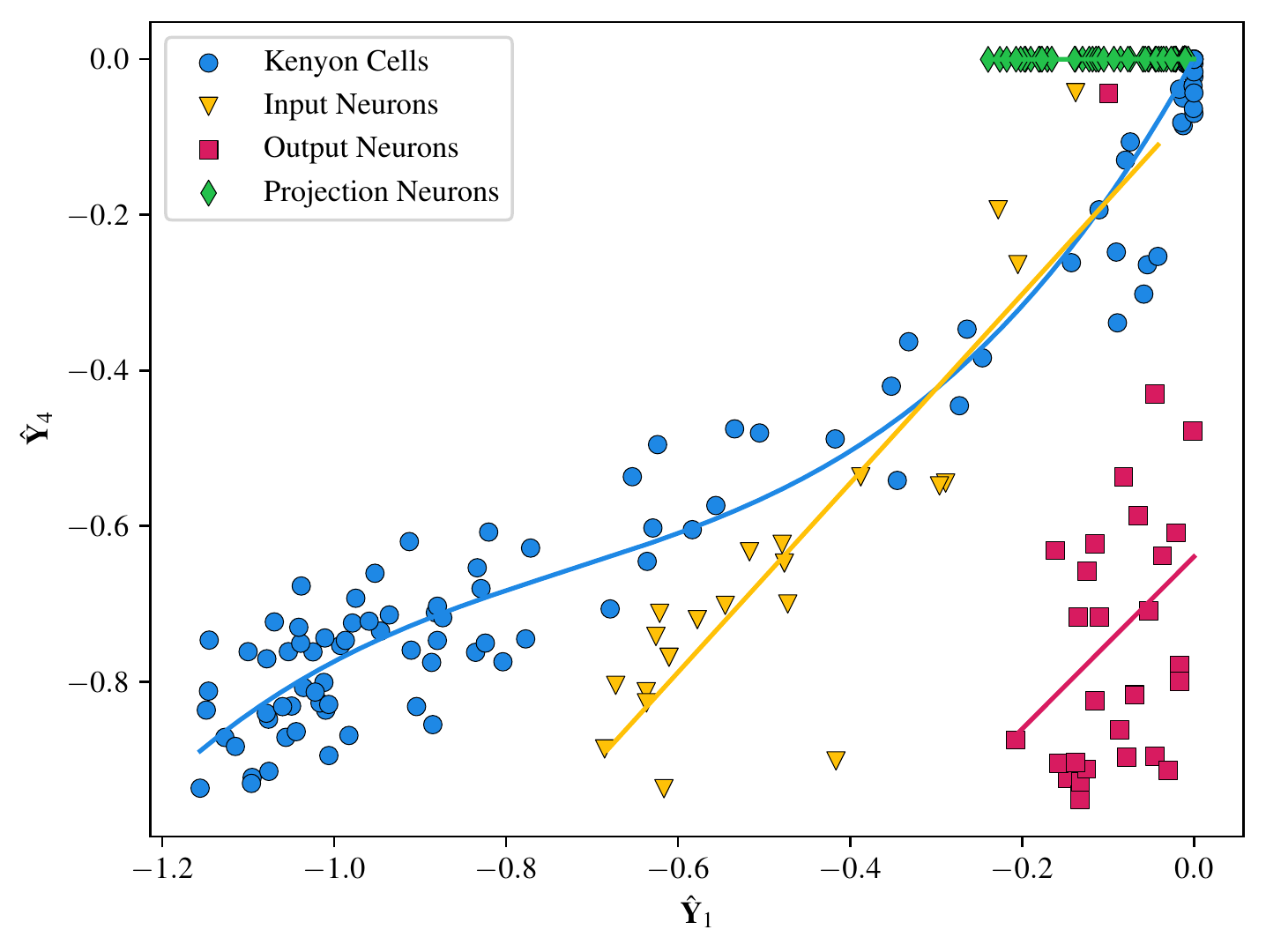}
\end{subfigure}
\begin{subfigure}[t]{0.325\textwidth}
\centering
\caption{$\hat{\mvec Y}_5$ vs. $\hat{\mvec Y}_1$}
\includegraphics[width=\textwidth]{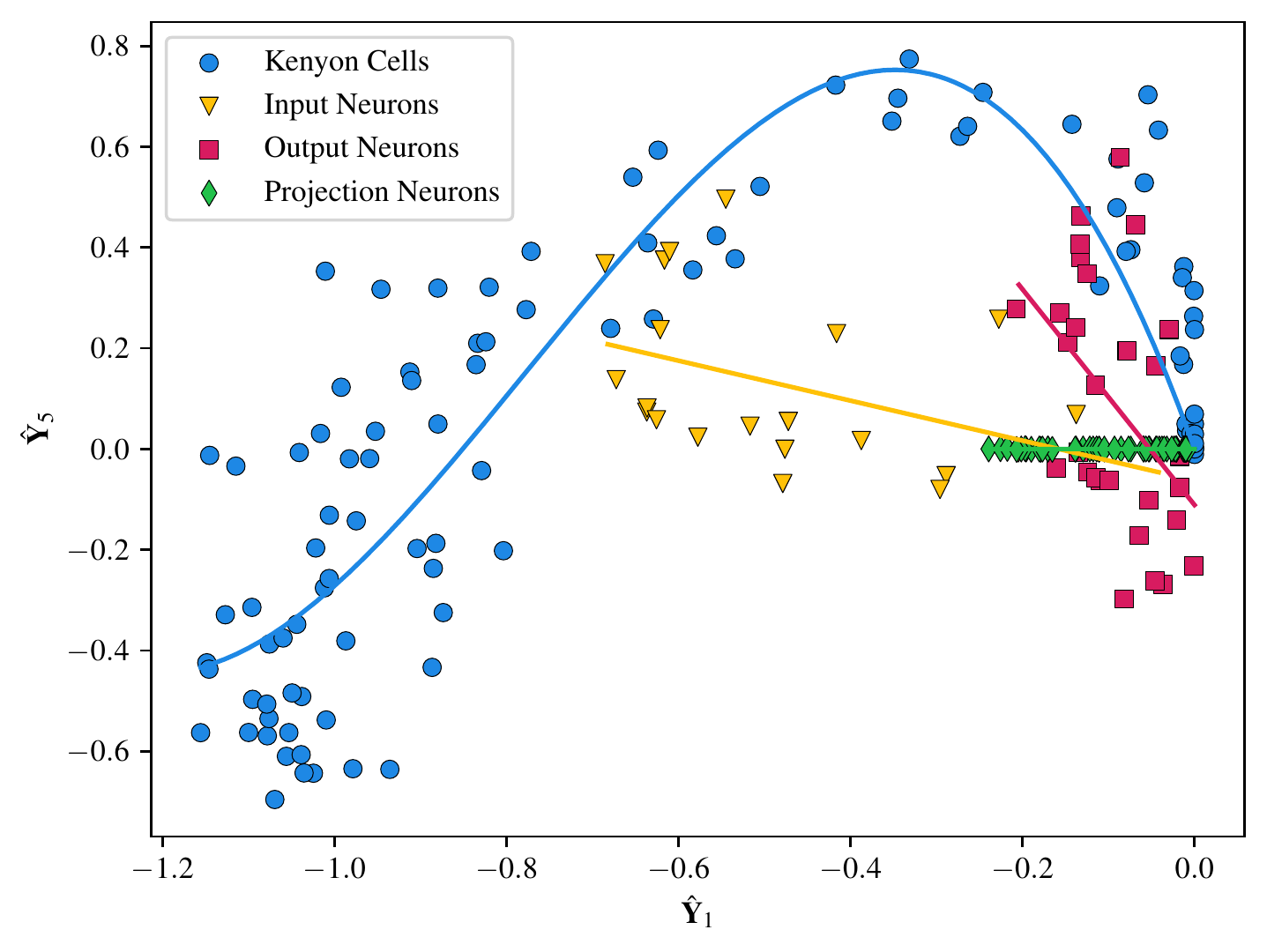}
\end{subfigure}
\begin{subfigure}[t]{0.325\textwidth}
\centering
\caption{$\hat{\mvec Y}_6$ vs. $\hat{\mvec Y}_1$}
\includegraphics[width=\textwidth]{Pictures/ddroso_15.pdf}
\end{subfigure}
\caption{Scatterplots of $\{\hat{\mvec Y}_2,\hat{\mvec Y}_3,\hat{\mvec Y}_4,\hat{\mvec Y}_5,\hat{\mvec Y}_6\}$ vs. $\hat{\mvec Y}_1$, coloured by neuron type, and best fitting latent functions obtained from the estimated clustering, plotted only over the range of nodes assigned to each community.}
\label{scatters_droso_full}
\end{figure*}

From the scatterplots in Figure~\ref{scatters_droso}, it appears that the embedding for the Input Neurons, Output Neurons and Projection Neurons could be also represented using linear functions. Furthermore, a quadratic curve for the Kenyon Cells might be too restrictive. The LSBM framework allows specification of all such choices. Here, assuming $f_{k,1}(\theta)=\theta$ for each $k$, the first community (Kenyon Cells) is given a cubic latent function, implying $\xi_{1,j}(\theta,\theta^\prime)=(\theta,\theta^2,\theta^3)\bm\Delta_{1,j}(\theta^\prime,\theta^{\prime2},\theta^{\prime3})^\intercal$. For the second community (Input Neurons), a latent linear function is used: $\xi_{2,j}(\theta,\theta^\prime)=(1,\theta)\bm\Delta_{2,j}(1,\theta^\prime)^\intercal,\ j=2,\dots,6$. Similarly, from observation of the scatterplots in Figure~\ref{scatters_droso}, the following kernels, corresponding to linear latent functions, are assigned to the remaining communities: $\xi_{3,j}(\theta,\theta^\prime)=\theta\theta^\prime\Delta_{3,j},\ j=2,\dots,6$, $\xi_{4,2}(\theta,\theta^\prime)=\theta\theta^\prime\Delta_{4,2}$, and $\xi_{4,j}(\theta,\theta^\prime)=(1,\theta)\bm\Delta_{k,j}(1,\theta^\prime)^\intercal,\ j=3,\dots,6$. The results for these choices of covariance kernels are plotted in Figure~\ref{scatters_droso_full}, resulting in ARI $0.8754$ for the estimated clustering, again corresponding to $10$ misclassified nodes. Note that this representation seems to capture more closely the structure of the embeddings. 

The performance of the LSBM is also compared to alternative methods for clustering in Table~\ref{clust_table_droso}. In particular, Gaussian mixture models with $K=4$ components were fitted on the concatenated DASE embedding $\hat{\mvec Y}=[\hat{\mvec X},\hat{\mvec X}^\prime]$ \citep[standard spectral clustering; see, for example][]{RubinDelanchy17}, on its row-normalised version $\tilde{\mvec Y}$ \citep{Ng01,Qin13}, and on a transformation to spherical coordinates \citep{SannaPassino20}. 
The ARI is averaged over 1000 different initialisations. 
Furthermore, the LSBM is also compared to PGP \citep{Bouveyron15}, hierarchical Louvain 
adapted to directed graphs \citep{Dugue15}, and hierarchical clustering 
on $\hat{\mvec Y}$. 
The results in Table~\ref{clust_table_droso} show that the LSBM largely outperform the alternative clustering techniques, which are not able to account for the non-linearity in the community corresponding to the Kenyon Cells.

\begin{table*}[t]
\centering
\scalebox{0.85}{
\begin{tabular}{c | cccccccc}
\toprule
Method & LSBM($\hat{\mvec Y}$) & LSBM($\hat{\mvec Y}$) & GMM($\hat{\mvec Y}$) & GMM($\tilde{\mvec Y}$) & SCSC($\hat{\mvec Y}$) & PGP($\hat{\mvec Y}$) & HLouvain & HClust($\hat{\mvec Y}$) \\
& Polynomial & Quadratic + SBM \\
\midrule
ARI & 0.875 & 0.864 & 0.599 & 0.585 & 0.667 & 0.555 & 0.087 & 0.321  \\
\bottomrule
\end{tabular}
}
\caption{ARI for communities estimated via LSBM and alternative methodologies on the \textit{Drosophila} connectome.}
\label{clust_table_droso}
\end{table*}

\begin{figure*}[t]
\centering
\begin{subfigure}[t]{0.49\textwidth}
\centering
\caption{Prior on kernels: Quadratic + SBM}
\label{droso_priebe_Khat}
\includegraphics[width=\textwidth]{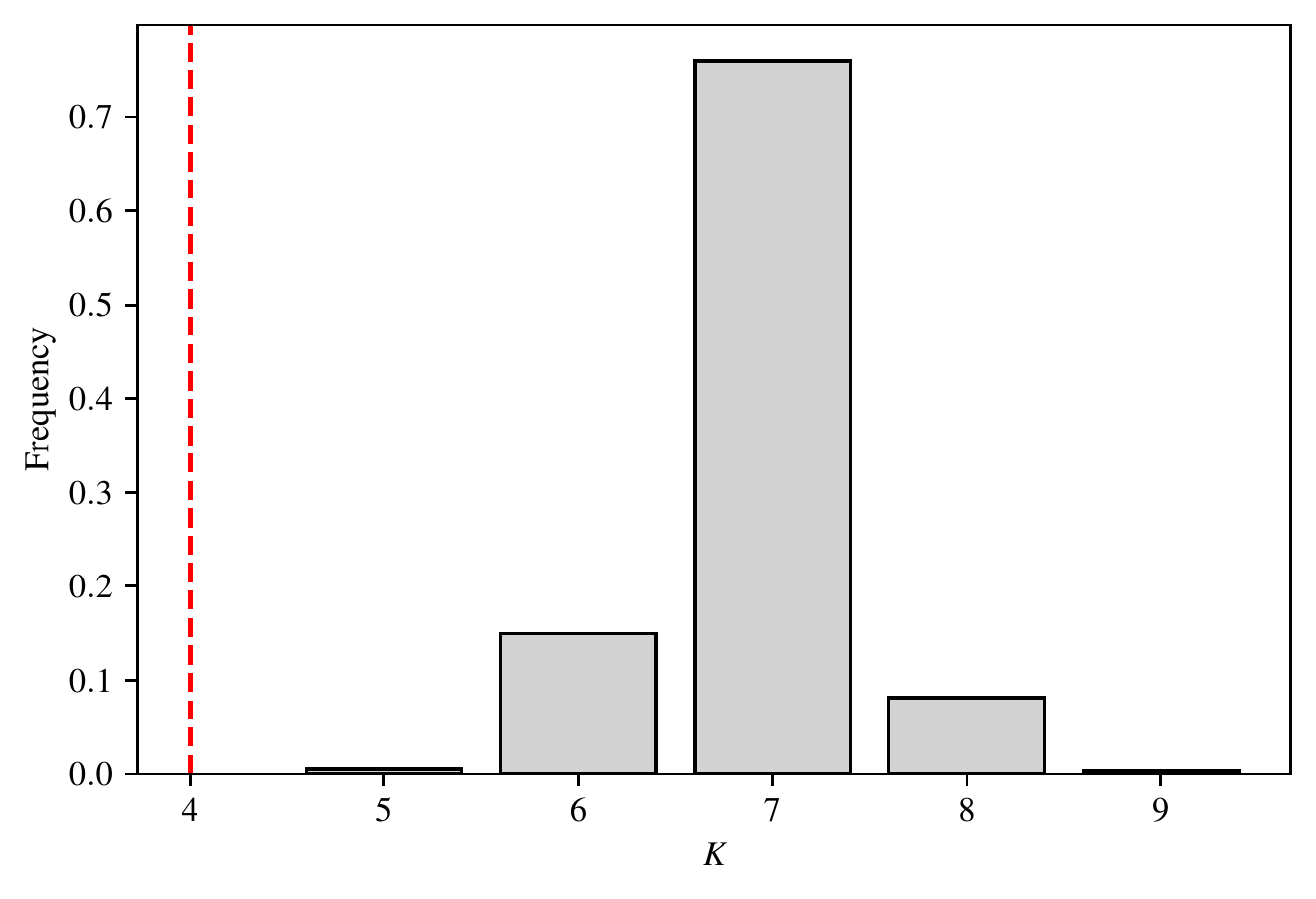}
\end{subfigure}
\begin{subfigure}[t]{0.49\textwidth}
\centering
\caption{Prior on kernels: Cubic + Linear}
\label{droso_full_Khat}
\includegraphics[width=\textwidth]{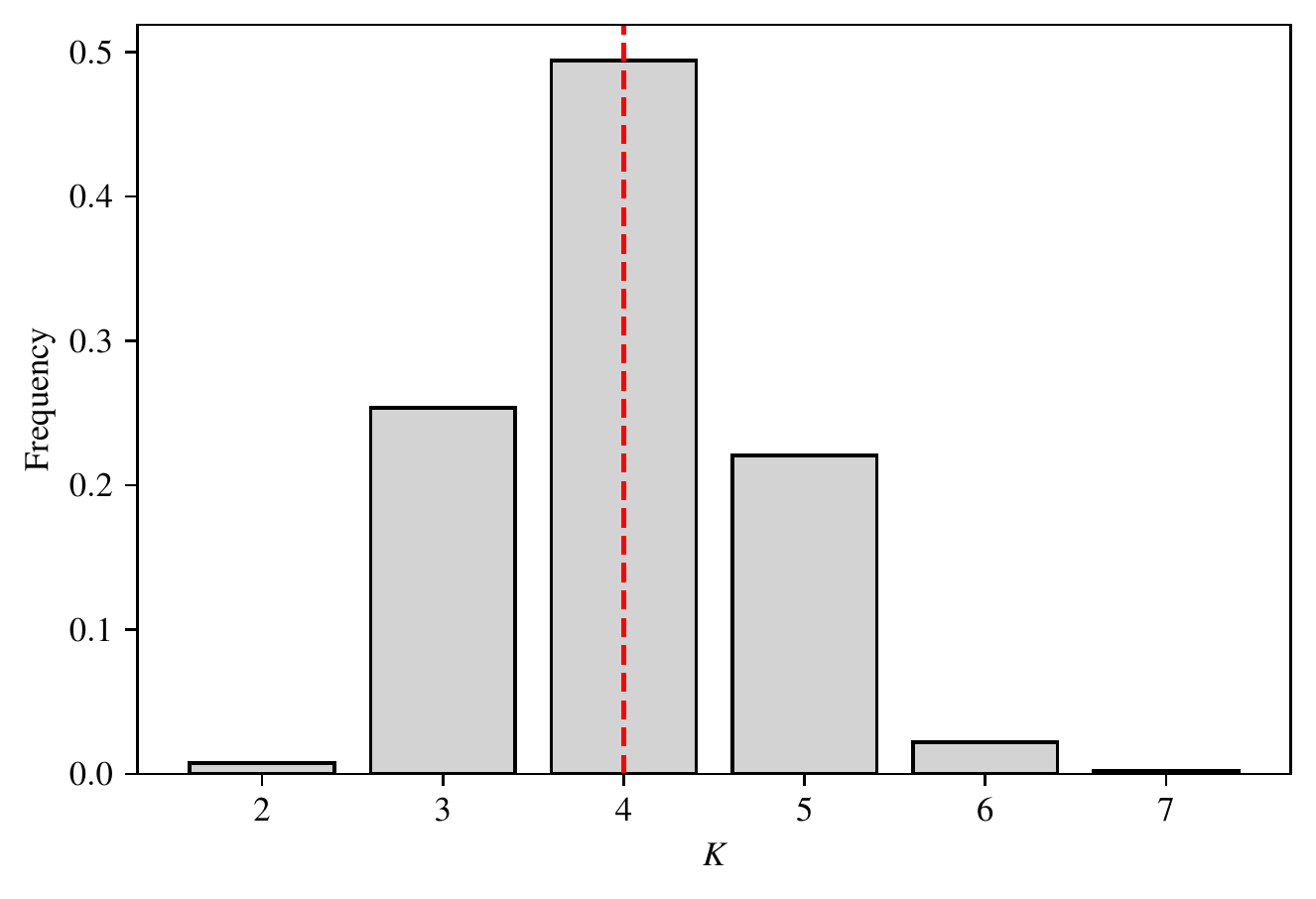}
\end{subfigure}
\caption{Estimated marginal posterior for the number of non-empty clusters $K$ on the \textit{Drosophila} connectome, under two different mixture prior distributions on the kernels.}
\label{scatters_droso_K}
\end{figure*}

Next, the MCMC algorithm is run for $M=\numprint{50000}$ iterations with $\numprint{5000}$ burn-in period, with the objective of inferring $K$, assuming a discrete uniform prior on the kernels used in Figure~\ref{scatters_droso} and \ref{scatters_droso_full}. The estimated posterior barplots for the number of non-empty clusters $K$ are plotted in Figure~\ref{scatters_droso_K}. It appears that the choice of a prior on kernels admitting a quadratic curve for the Kenyon Cells and Gaussian clusters for the remaining groups is overly simplistic, since some of the communities appear to have a linear structure (\textit{cf.} Figure~\ref{scatters_droso}). Hence, the number of communities $K$ is incorrectly estimated  under this setting (\textit{cf.} Figure~\ref{droso_priebe_Khat}). On the other hand, when a mixture distribution of cubic and linear kernels is used as prior on $\bm\xi_k$, $K$ is correctly estimated (\textit{cf.} Figure~\ref{droso_full_Khat}), confirming the impression from Figure~\ref{scatters_droso_full} of a better fit of such kernels on the \textit{Drosophila} connectome. 

\subsection{Bipartite graphs: ICL computer laboratories} \label{sec:icl}

\begin{figure*}[t]
\centering
\begin{subfigure}[t]{0.49\textwidth}
\centering
\caption{$\hat{\mvec X}_2$ vs. $\hat{\mvec X}_1$}
\includegraphics[width=\textwidth]{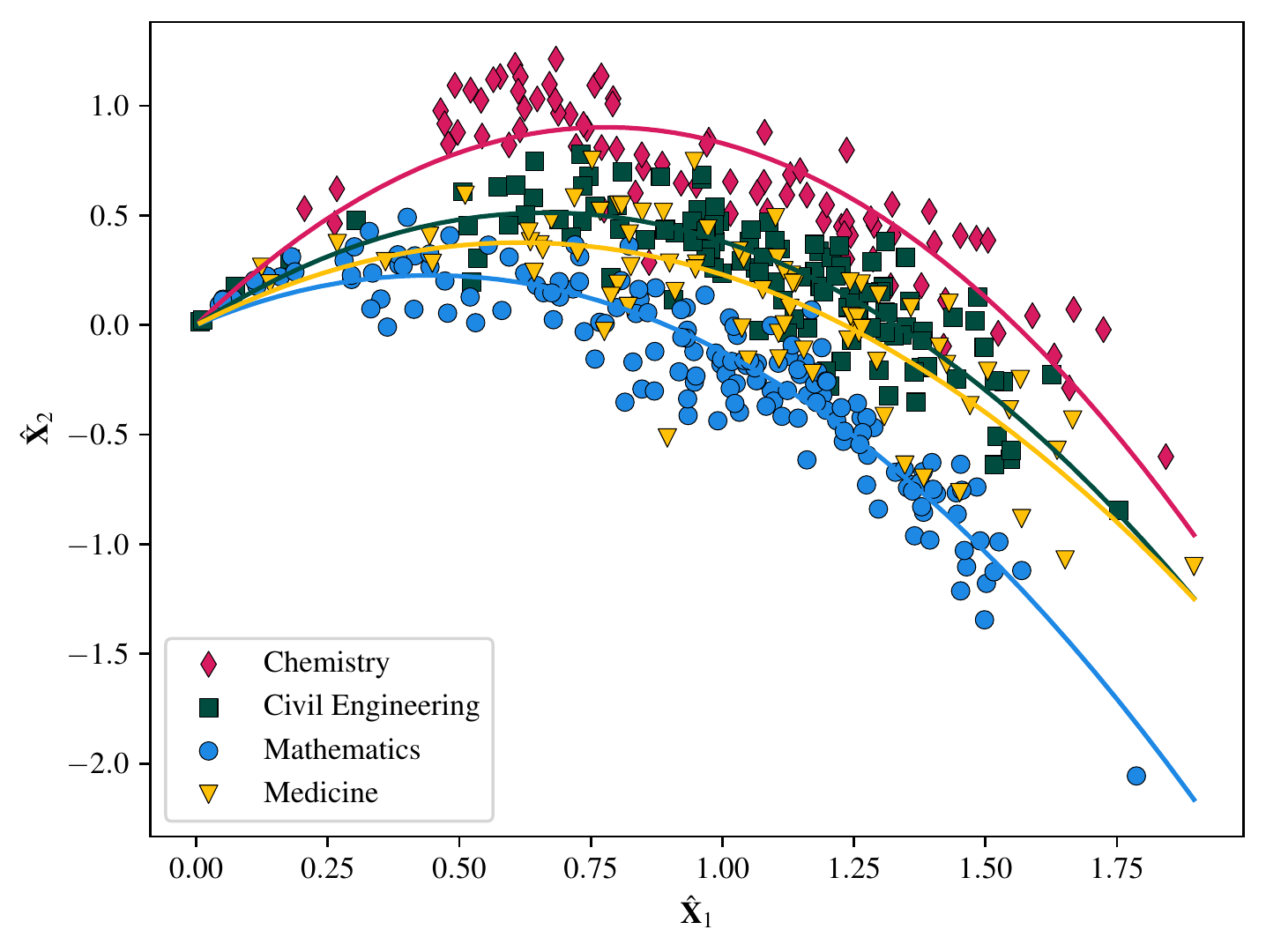}
\end{subfigure}
\begin{subfigure}[t]{0.49\textwidth}
\centering
\caption{$\hat{\mvec X}_3$ vs. $\hat{\mvec X}_1$}
\includegraphics[width=\textwidth]{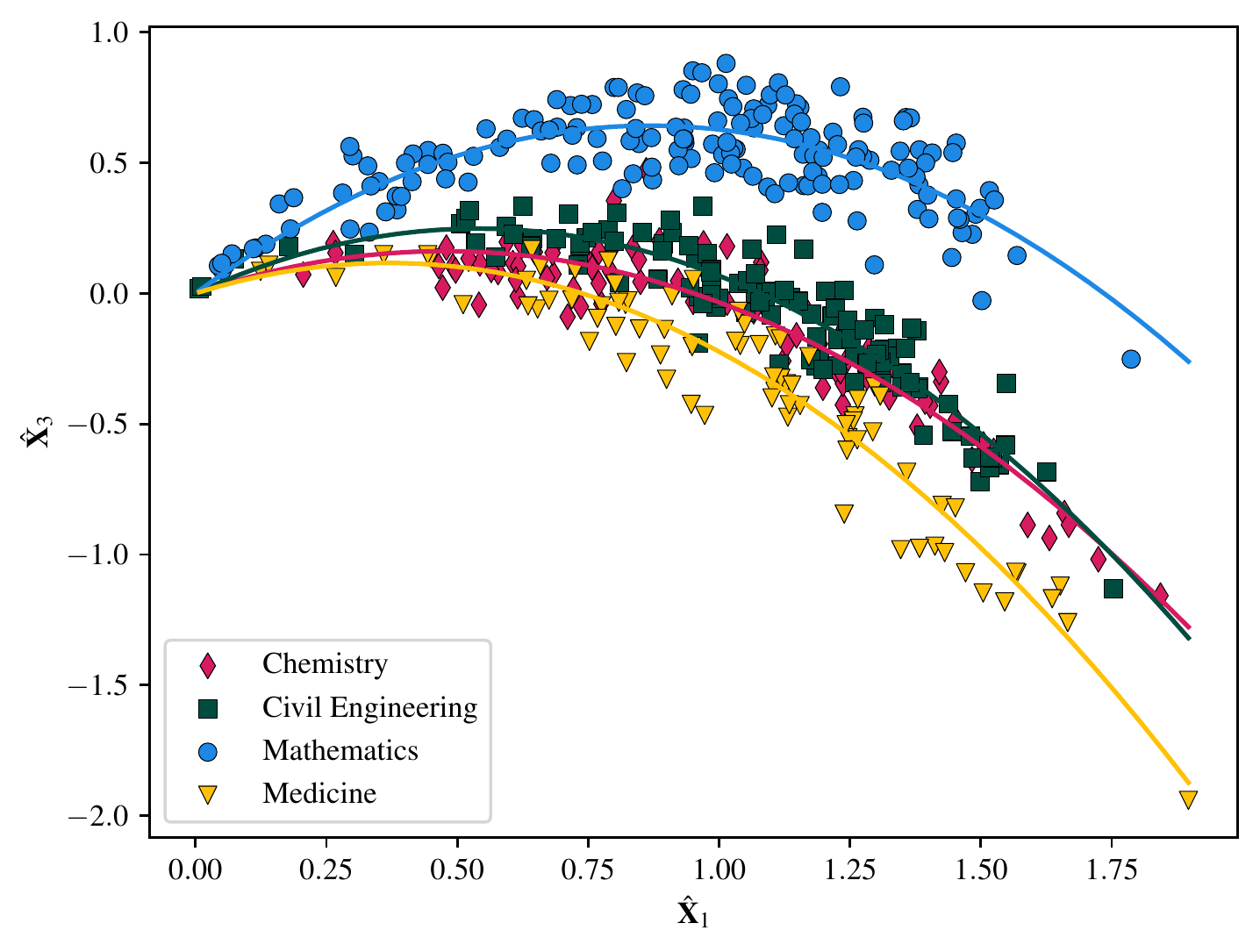}
\end{subfigure}
\centering
\begin{subfigure}[t]{0.49\textwidth}
\centering
\caption{$\hat{\mvec X}_4$ vs. $\hat{\mvec X}_1$}
\includegraphics[width=\textwidth]{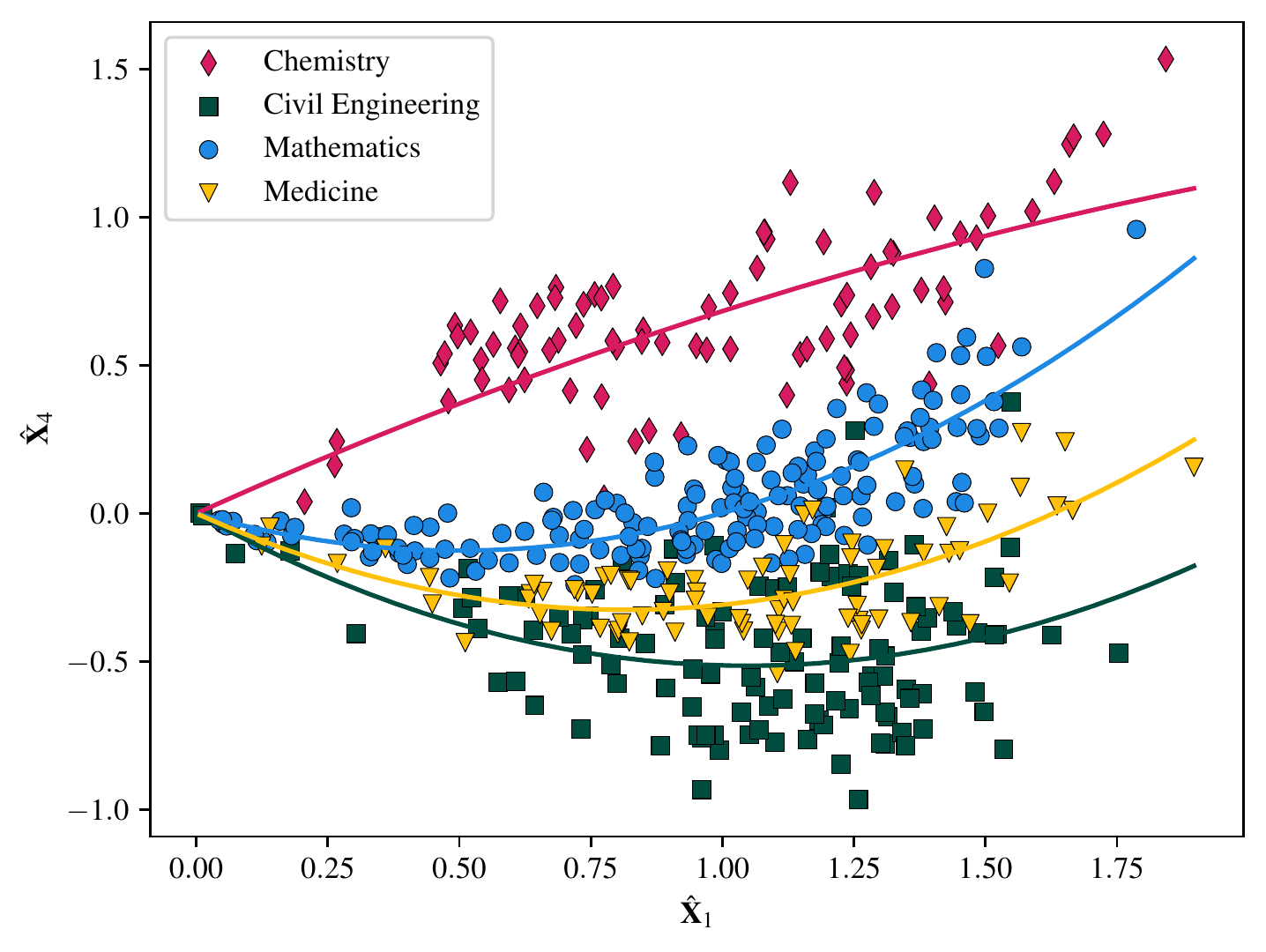}
\end{subfigure}
\begin{subfigure}[t]{0.49\textwidth}
\centering
\caption{$\hat{\mvec X}_5$ vs. $\hat{\mvec X}_1$}
\includegraphics[width=\textwidth]{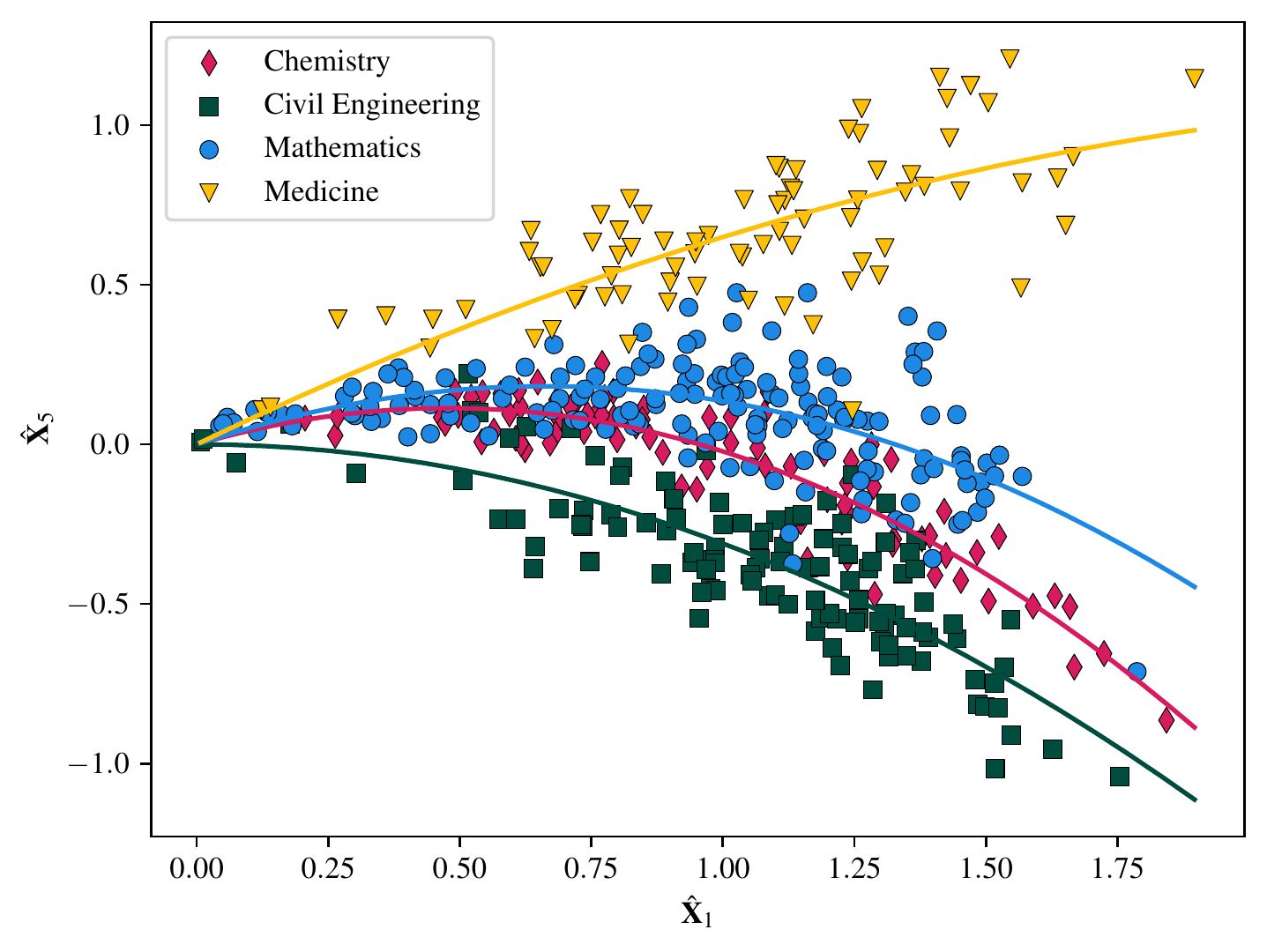}
\end{subfigure}
\caption{Scatterplots of $\{\hat{\mvec X}_2,\hat{\mvec X}_3,\hat{\mvec X}_4,\hat{\mvec X}_5\}$ vs. $\hat{\mvec X}_1$, coloured by department, and best fitting quadratic curves passing through the origin.}
\label{scatters}
\end{figure*}

\begin{figure}[t]
\centering
\includegraphics[width=0.475\textwidth]{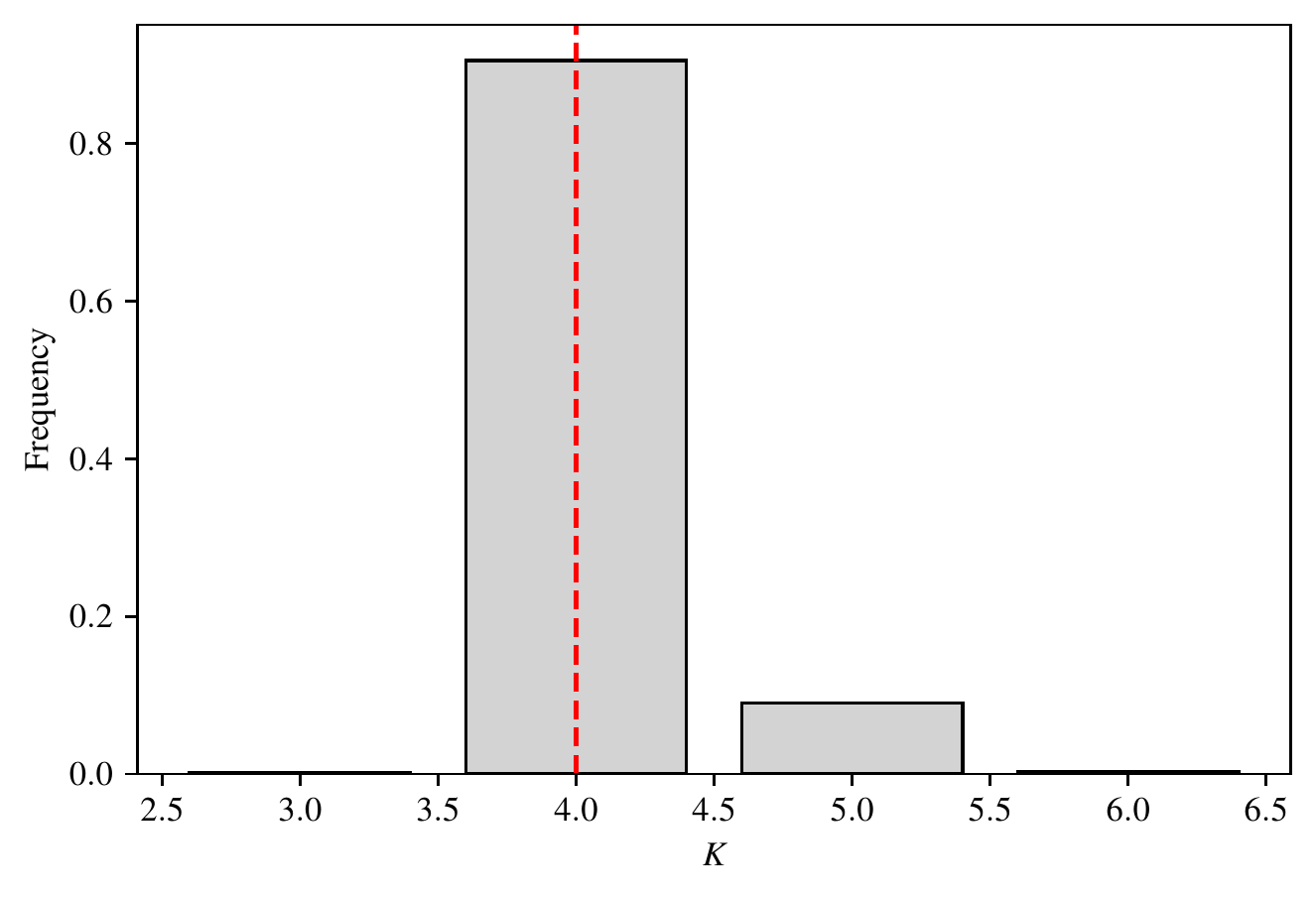}
\caption{Estimated marginal posterior for the number of non-empty clusters $K$ on the ICL NetFlow network, under quadratic kernels.}
\label{icl_full_Khat}
\end{figure}

The LSBM methodology is finally applied to a bipartite graph obtained from computer network flow data collected at Imperial College London (ICL). The source nodes are $\abs{V_1}=n=439$ client machines located in four computer laboratories in different departments at Imperial College London, whereas the destination nodes are $\abs{V_2}=\numprint{60635}$ internet servers, connected to by HTTP and HTTPS in January 2020 from one or more of the $439$ client computers. A total of $\numprint{717912}$ edges are observed. The inferential objective is to identify the location of the machines in the network, represented by their department, from the realisation of the rectangular adjacency matrix $\mvec A$, where $A_{ij}=1$ if at least one connection is observed between client computer $i\in V_1$ and the server $j\in V_2$, and $A_{ij}=0$ otherwise. It could be assumed that $K=4$, representing the departments of Chemistry, Civil Engineering, Mathematics, and Medicine. After taking the DASE of $\mvec A$, the machines are clustered using the LSBM. The value $d=5$ is selected using the criterion of \citet{zhu}, choosing the second elbow of the scree-plot of singular values. 
Computer network graphs of this kind have been seen to present quadratic curves in the embedding, as demonstrated, for example, by Figure~\ref{maths_cive} in the introduction, which refers to a different set of machines. 
Therefore, it seems reasonable to assume that 
$\xi_{k,j}(\theta,\theta^\prime)=(\theta,\theta^2)\bm\Delta_{k,j}(\theta^\prime,\theta^{\prime2})^\intercal,\ j=2,\dots,d$,
which implies quadratic functions passing through the origin, and $f_{k,1}(\theta)=\theta$.
The quadratic model with $K=4$ is fitted to the $5$-dimensional embedding, obtaining ARI $0.9402$, corresponding to just $9$ misclassified nodes. The results are plotted in Figure~\ref{scatters}, with the corresponding best fitting quadratic curves obtained from the estimated clustering. The result appears remarkable, considering that the communities are highly overlapping. Running the MCMC algorithm assuming $K$ unknown shows that the number of clusters is correctly estimated (\textit{cf.} Figure~\ref{icl_full_Khat}), overwhelmingly suggesting $K=4$, corresponding to the correct number of departments. 

These LSBM results are also compared to alternative methodologies in Table~\ref{clust_table}. LSBM achieves the best performance in clustering the nodes, followed by Gaussian mixture modelling of the row-normalised adjacency spectral embedding. The GMM on $\tilde{\mvec X}$ sometimes converges to competitive solutions, reaching ARI up to $0.94$, but usually converges to sub-optimal solutions, as demonstrated by the average ARI of $0.7608$.

\begin{table*}[t]
\centering
\scalebox{0.85}{
\begin{tabular}{c | cccccccc}
\toprule
Method & LSBM($\hat{\mvec X}$) & LSBM($\hat{\mvec X}$) & GMM($\hat{\mvec X}$) & GMM($\tilde{\mvec X}$) & SCSC($\hat{\mvec X}$) & PGP($\hat{\mvec X}$) & HLouvain & HClust($\hat{\mvec X}$) \\
& Quadratic & Splines \\
\midrule
ARI & 0.940 & 0.936 & 0.659 & 0.766 & 0.921 & 0.895 & 0.602 & 0.139 \\
\bottomrule
\end{tabular}
}
\caption{ARI for communities estimated via LSBM and alternative methodologies on the ICL NetFlow network.}
\label{clust_table}
\end{table*}

\begin{figure*}[t]
\centering
\begin{subfigure}[t]{0.49\textwidth}
\centering
\caption{$\hat{\mvec X}_2$ vs. $\hat{\mvec X}_1$}
\includegraphics[width=\textwidth]{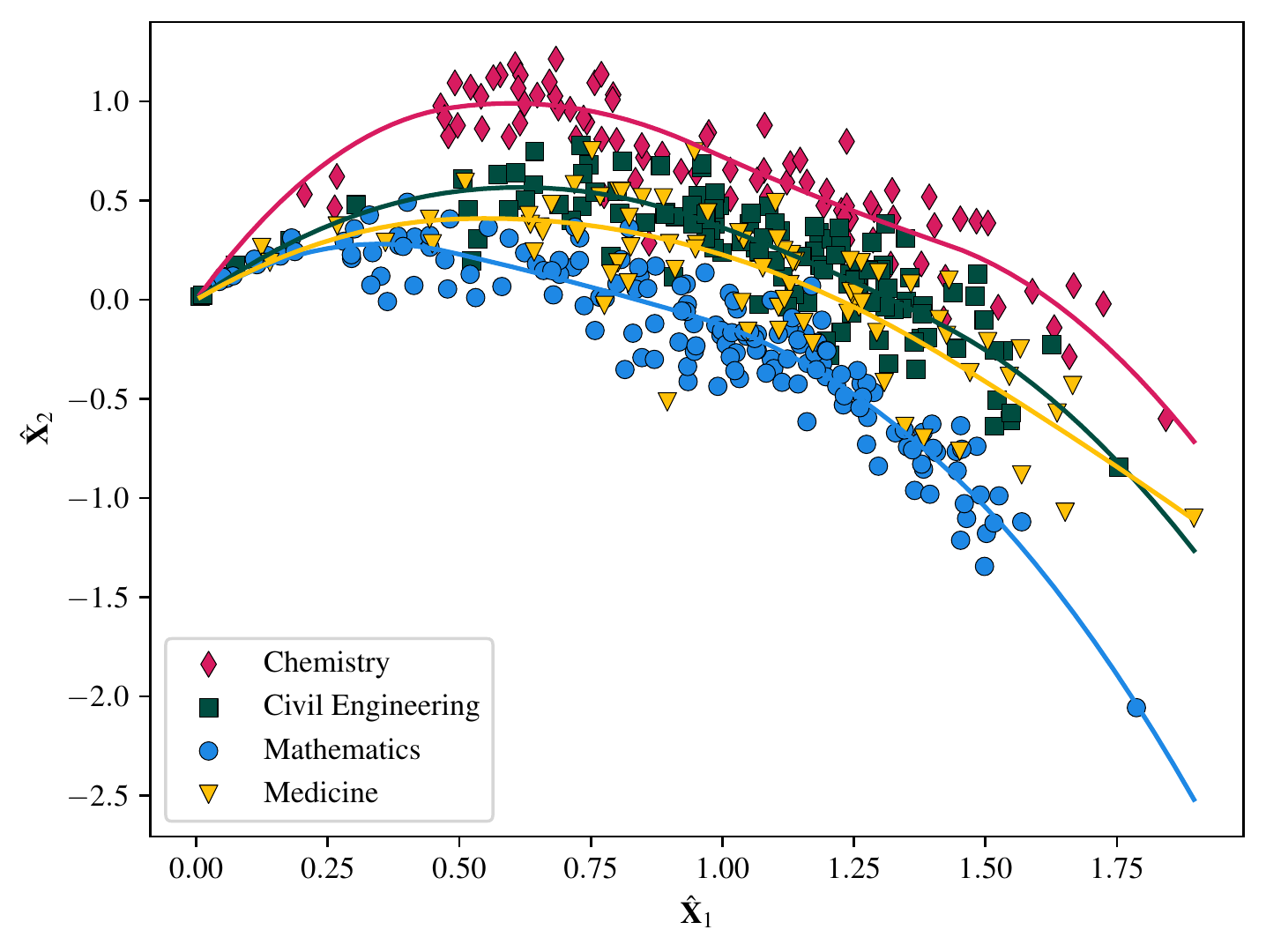}
\end{subfigure}
\begin{subfigure}[t]{0.49\textwidth}
\centering
\caption{$\hat{\mvec X}_3$ vs. $\hat{\mvec X}_1$}
\includegraphics[width=\textwidth]{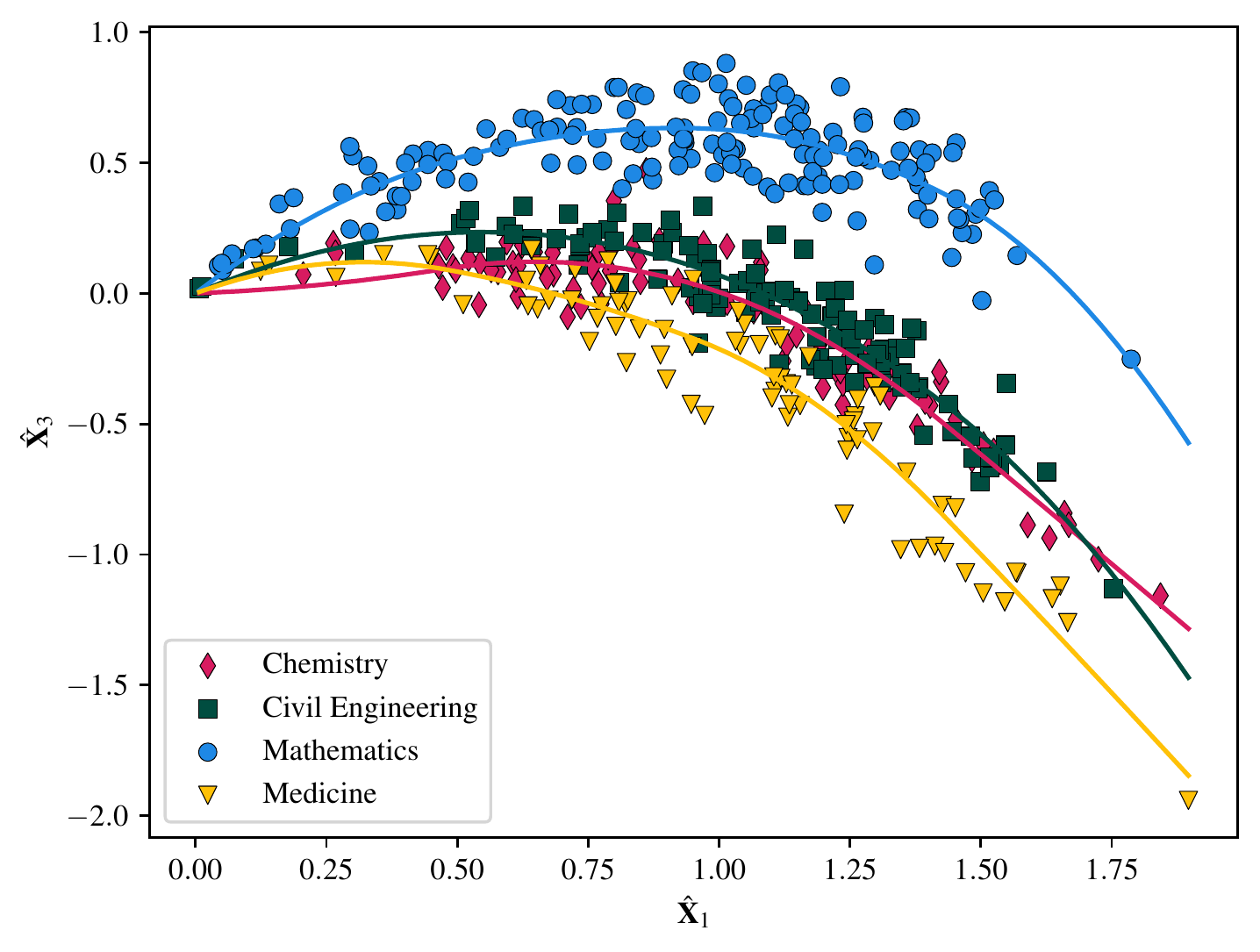}
\end{subfigure}
\begin{subfigure}[t]{0.49\textwidth}
\centering
\caption{$\hat{\mvec X}_4$ vs. $\hat{\mvec X}_1$}
\includegraphics[width=\textwidth]{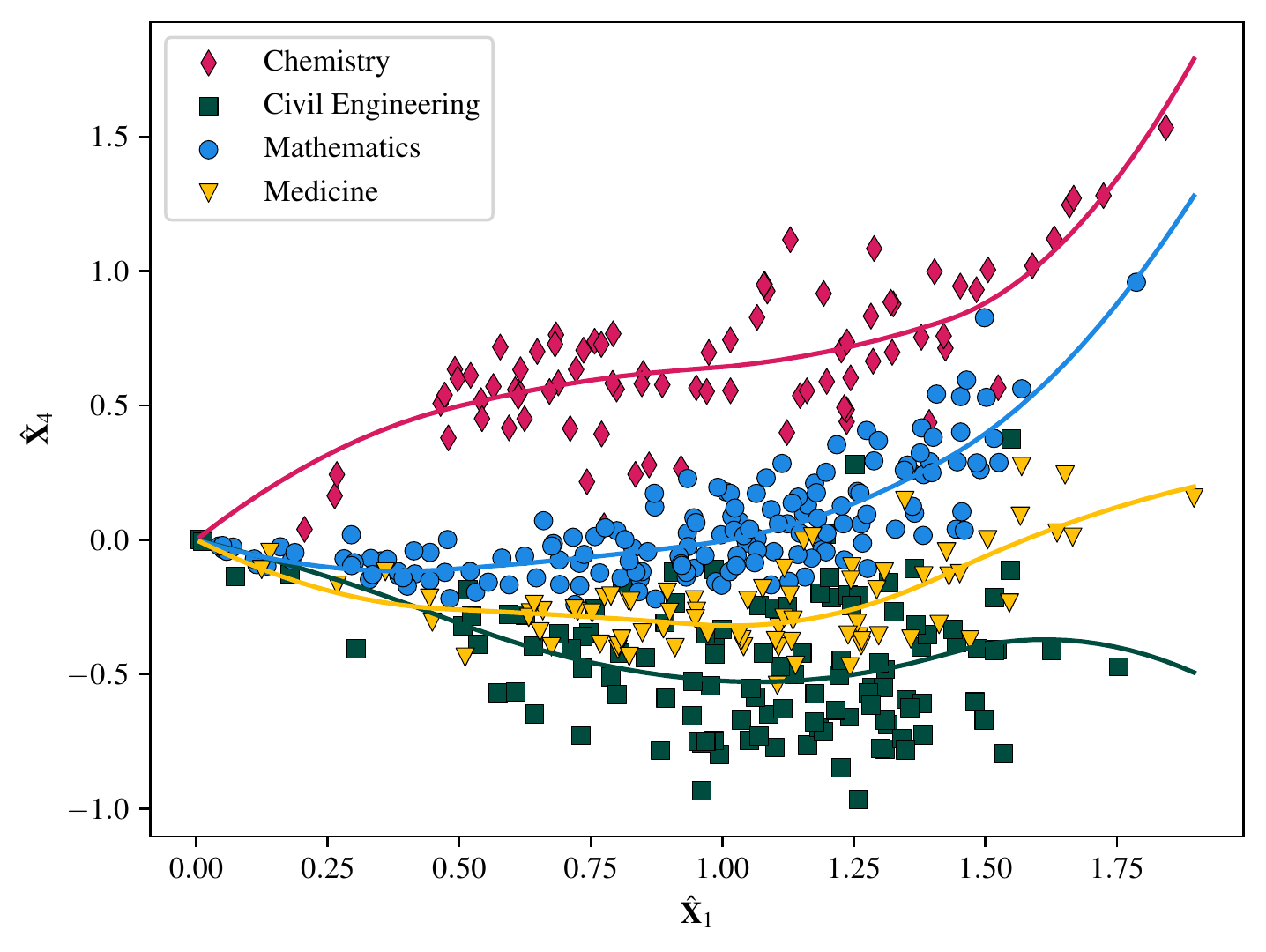}
\end{subfigure}
\begin{subfigure}[t]{0.49\textwidth}
\centering
\caption{$\hat{\mvec X}_5$ vs. $\hat{\mvec X}_1$}
\includegraphics[width=\textwidth]{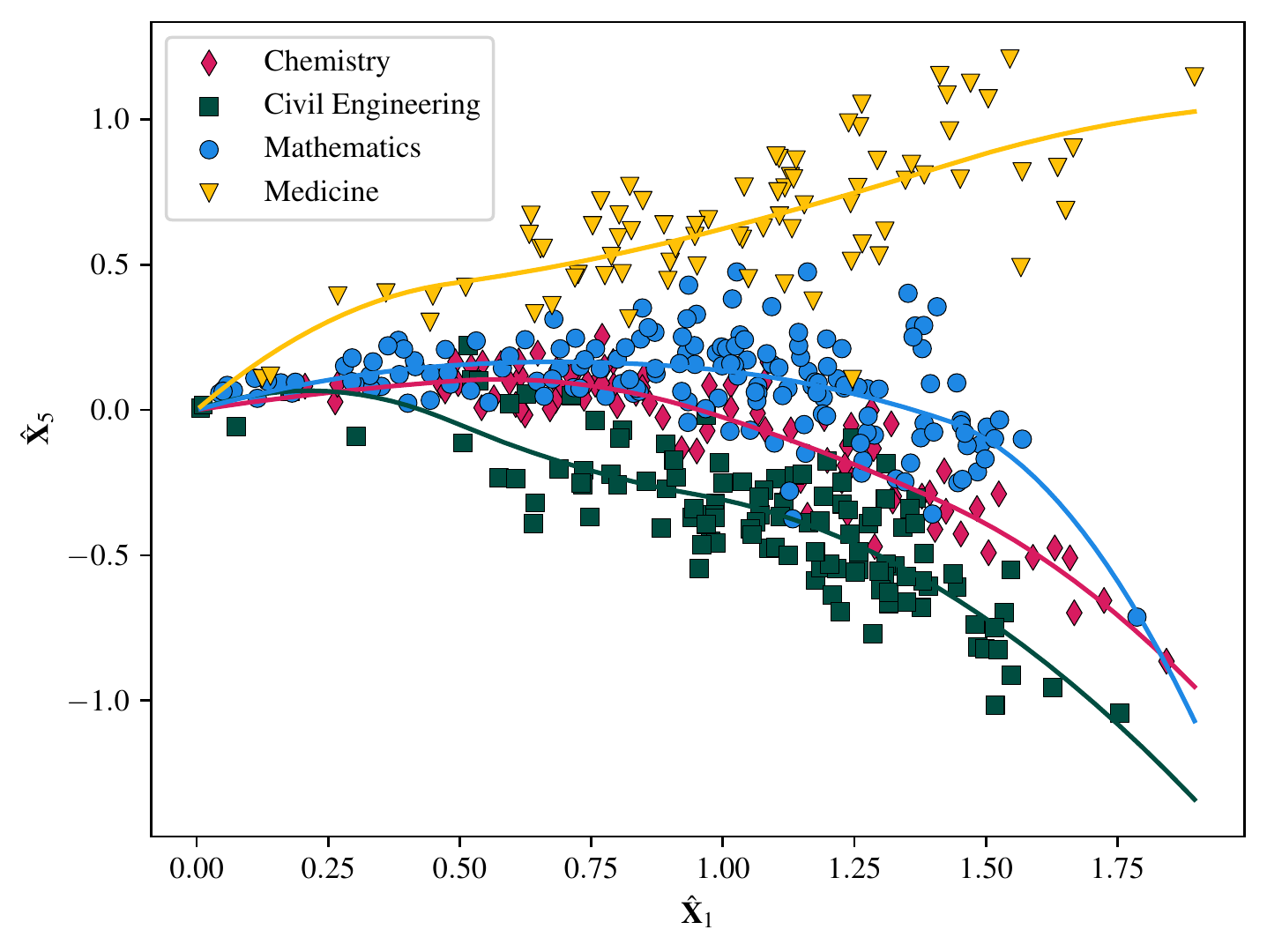}
\end{subfigure}
\caption{Scatterplots of $\{\hat{\mvec X}_2,\hat{\mvec X}_3,\hat{\mvec X}_4,\hat{\mvec X}_5\}$ vs. $\hat{\mvec X}_1$, coloured by department, and estimated cubic truncated power splines passing through the origin.}
\label{scatter_splines}
\end{figure*}

As before, if a parametric form for $\vec f_k(\cdot)$ is unknown, regression splines can be used, for example the truncated power basis \eqref{spline_functions}, with three equispaced knots in the range of $\hat{\mvec X}_1$.
The results for the initial three dimensions are plotted in Figure~\ref{scatter_splines}. The communities are recovered correctly, and the ARI is $0.9360$, corresponding to $10$ misclassified nodes. 

\section{Conclusion} \label{sec:conclusion}

An extension of the latent structure model \citep{Athreya21} for networks has been introduced, and inferential procedures based on Bayesian modelling of spectrally embedded nodes have been proposed. The model, referred to as the latent structure blockmodel (LSBM), allows for latent positions living on community-specific univariate structural support manifolds, using flexible Gaussian process priors. Under the Bayesian paradigm, most model parameters can be integrated out and inference can be performed efficiently. The proposed modelling framework could be utilised also when the number of communities is unknown and there is uncertainty around the choice the Gaussian process kernels, encoded by a prior distribution. The performance of the model inference has been evaluated on simulated and real world networks. In particular, excellent results have been obtained on complex clustering tasks concerning the \textit{Drosophila} connectome and the Imperial College NetFlow data, where a substantial overlap between communities is observed. Despite these challenges, the proposed methodology is still able to recover a correct clustering. 

Overall, this work provides a modelling framework for graph embeddings arising from random dot product graphs where it is suspected that nodes belong to community-specific lower-dimensional subspaces. In particular, this article discusses the case of curves, which are one-dimensional structural support submanifolds. The methodology has been demonstrated to have the potential to recover the correct clustering structure even if the underlying parametric form of the underlying structure is unknown, using a flexible Gaussian process prior on the unknown functions. In particular, regression splines with a truncated power basis have been used, showing good performance in recovering the underlying curves. 


In the 
model proposed in this work, it has been assumed that the variance only depends on the community allocation. This enables marginalising most of the parameters leading to efficient inference. On the other hand, this is potentially an oversimplification, since the ASE-CLT 
(Theorem~\ref{clt}) establishes that the covariance structure depends on the latent position. Further work should study efficient algorithms for estimating the parameters when an explicit functional form dependent on $\theta_i$ is incorporated in the covariance.

\section*{Code}

A \textit{python} library to reproduce the results in this paper is available in the \textit{GitHub} repository \href{https://www.github.com/fraspass/lsbm}{\texttt{fraspass/lsbm}}.

\section*{Acknowledgements}

FSP and NAH gratefully acknowledge funding from the Microsoft Security AI research grant \textit{"Understanding the enterprise: Host-based event prediction for automatic defence in  cyber-security"}.


\bibliographystyle{rss2}
\singlespacing
\bibliography{biblio}

\appendix

\end{document}